\documentclass{article}

\usepackage[nonatbib,final]{neurips_2022}

\usepackage[utf8]{inputenc} 
\usepackage[T1]{fontenc}    
\usepackage[backref=page]{hyperref}       
\usepackage{url}            
\usepackage{booktabs}       
\usepackage{amsfonts}       
\usepackage{nicefrac}       
\usepackage{microtype}      
\usepackage{xcolor}         

\usepackage{amsmath,amssymb} 
\usepackage{graphicx}
\usepackage{multirow}
\usepackage{caption}
\newcommand\revise[1]{\textcolor{black}{#1}}
\title{Local Spatiotemporal Representation Learning for Longitudinally-consistent Neuroimage Analysis}

\author{%
  Mengwei Ren\\
  New York University\\
  \texttt{mengwei.ren@nyu.edu} \\
  \And
  Neel Dey \\
  New York University\\
  \texttt{neel.dey@nyu.edu} \\
  \And
  Martin A. Styner \\
  UNC-Chapel Hill \\
  \texttt{styner@cs.unc.edu} \\
  \AND 
  Kelly N. Botteron\\
  WUSTL School of Medicine \\
  \texttt{botteronk@wustl.edu}
  \And
  Guido Gerig \\
  New York University\\
  \texttt{gerig@nyu.edu} \\
}
\begin{document}

\maketitle

\begin{abstract}
Recent self-supervised advances in medical computer vision exploit the global and local \revise{anatomical self-similarity} for pretraining prior to downstream tasks such as segmentation. However, current methods assume i.i.d. image acquisition, which is invalid in \revise{clinical study designs} where follow-up \textit{longitudinal} scans track subject-specific temporal changes. Further, existing self-supervised methods for medically-relevant image-to-image architectures exploit only spatial or temporal self-similarity and do so via a loss applied only at a single image-scale, with naive \textit{multi-scale spatiotemporal} extensions collapsing to degenerate solutions. To these ends, this paper \revise{makes two contributions}: (1) It presents a \textit{local} and \textit{multi-scale} spatiotemporal representation learning method for image-to-image architectures trained on longitudinal images. It exploits the spatiotemporal self-similarity of \revise{learned} multi-scale intra-subject image features for pretraining and develops several feature-wise regularizations that avoid \revise{degenerate} representations; (2) During finetuning, it proposes a surprisingly simple self-supervised segmentation consistency regularization to exploit intra-subject correlation. \revise{Benchmarked across various segmentation tasks}, the proposed framework outperforms both well-tuned randomly-initialized baselines and current self-supervised techniques designed for both i.i.d. and longitudinal datasets. These improvements are demonstrated across both longitudinal neurodegenerative adult MRI and developing infant brain MRI and yield both higher performance and longitudinal consistency. 
\end{abstract}

\section{Introduction}
\label{sec:introduction}
Tracking \textit{subject-specific} anatomical trends over time is crucial to both clinical diagnostics and large-scale biomedical science. Such \textit{longitudinal imaging} is especially relevant to analyzing neurological patterns of growth and degeneration via brain imaging in pediatric and elderly populations, respectively. As tracking individual structural changes requires precise and longitudinally-consistent segmentation methods with scarce annotated training volumes, we identify two major bottlenecks in existing self-supervised biomedical image analysis methods which we use to motivate our work.

\begin{figure}[t]
    \centering
    \includegraphics[width=\textwidth]{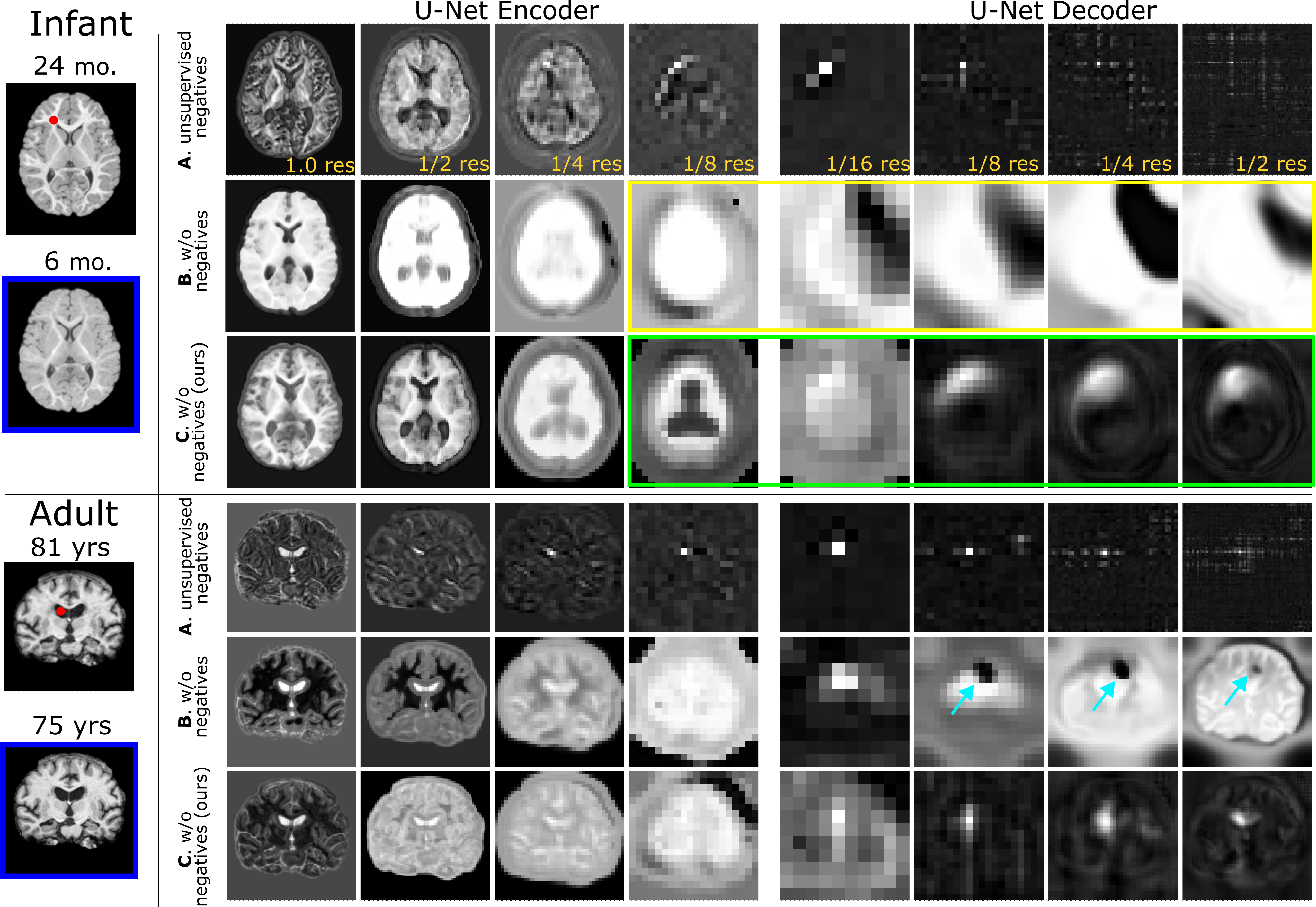}
    \caption{On pretraining an image-to-image network with \textit{per-layer} \revise{spatiotemporal} self-supervision, we visualize the \textbf{\revise{intra-subject multi-scale 
 feature similarity}} between a {\color{red} \textbf{query}} \revise{channel-wise feature and all spatial positions within the} {\color{blue} \textbf{key}} \revise{feature} at a different age. \textbf{A:} Contrastive pretraining with unsupervised negatives~\cite{park2020contrastive} yields only positionally-dependent representations. \textbf{B:} Pretraining w/o negatives~\cite{chen2020simple} by using corresponding intra-subject patch locations as positives leads to \revise{semantically-implausible representations with low-diversity (e.g., see yellow box)} and artifacts (\revise{see arrows}) in deeper layers. \textbf{C:} Our method attains both positionally and anatomically-relevant representations \revise{via proper regularization (e.g., see green box)}. \revise{Additional structures are visualized in Suppl. Figure~\ref{fig:simmap_more}.}}
    \label{fig:simmap}
\end{figure}

\textbf{Learning with few annotations.} While modern imaging studies may scan hundreds to thousands of individuals, manually outlining volumetric structures of interest across multiple individuals for supervised segmentation training is prohibitively expensive. Therefore, current work focuses on leveraging large sets of unlabeled images to \textit{pretrain} image-to-image architectures (e.g., the U-Net~\cite{ronneberger2015u}), which can then be efficiently finetuned in the one or few-shot setting. These \textit{self-supervised} methods may handcraft pre-text training objectives~\cite{chen2019self,gidaris2018unsupervised,pathak2016context,zhang2017split} or may attempt to pretrain the base network to be equivariant to transformations\revise{ in order to} preserve semantic meaning as in contrastive learning in medical~\cite{chaitanya2020contrastive,zeng2021positional} and natural image vision~\cite{Alonso_2021_ICCV,Hu_2021_ICCV,liu2022reco,Wang_2021_ICCV,Zhao_2021_ICCV,Zhong_2021_ICCV}. However, these methods typically leverage label supervision in sampling for their losses and impose a self-supervised loss only at a single feature scale (typically the encoder bottleneck or network output). Naive application of local unsupervised multi-scale contrastive losses~\cite{dey2022contrareg,park2020contrastive} lead to degenerate representations (Fig. \ref{fig:simmap}\revise{, row A}) when applied to both the encoder and decoder of image-to-image architectures.

\looseness=-1
\textbf{Violating i.i.d. assumptions.} Further, most existing self-supervised frameworks assume i.i.d. data\revise{. Unfortunately, this} assumption does not transfer to longitudinal studies where intra-subject temporal images are highly correlated.
Emerging longitudinal representation learning methods focus on imposing temporal-consistency into the encoder bottleneck~\cite{couronne2021longitudinal,ouyang2021self,zhao2021longitudinal}, such that the encoder learns representations that are aware of the order of acquisition~\cite{couronne2021longitudinal} and the overall trajectory~\cite{ouyang2021self}. These methods address image-level tasks such as disease classification or progression and age prediction. However, their extension to pixel-level applications with image-to-image architectures remains unclear.

\looseness=-1
\textbf{Methods.} Motivated by the above limitations, in this work, we claim that the spatiotemporal dependency of imaging data should be explicitly incorporated in self-supervised frameworks.
We do so by exploiting the spatial and temporal self-similarity of local multiscale deep features in both encoder and decoder and \revise{further learn diverse intermediate representations} by developing regularizations for self-supervised similarity objectives.
Lastly, when finetuning with limited annotated data, we encourage predictions on unlabeled subject-wise images to be spatiotemporally consistent.

\textbf{Contributions.} This work makes the following contributions: (1) It presents a longitudinally-consistent spatiotemporal representation learning framework to learn from image time-series; 
(2) To impose multi-scale local self-supervision \revise{while avoiding degenerate solutions,}
it develops regularization terms on the variance, covariance, orthogonality of local features within the decoder; (3) To further self-supervise the fine-tuning stage, the proposed method encourages segmentations from \revise{adjacent} timepoints on unlabeled data to be consistent; (4) Across three large-scale longitudinal \revise{one, few, and full-}shot segmentation tasks on both elderly and pediatric populations, the developed framework yields improved segmentation performance and higher longitudinal segmentation consistency. Our code is available at {\color{red}\url{https://www.mengweiren.com/research/spatiotemporal-learning/}}.

\section{Related work}
\label{sec:related_works}

\looseness=-1
\textbf{Self-supervision.}
Self-supervised learning (SSL) methods aim to learn hierarchical representations from unannotated data, which can then be transferred to tasks operating in low-annotation regimes. Early work focused on pretext tasks where a handcrafted loss is used to pretrain networks via orientation prediction~\cite{gidaris2018unsupervised}, context restoration~\cite{chen2019self,pathak2016context}, channel prediction~\cite{zhang2017split}, among others. Given their heuristic nature and suboptimal generalization, recent work instead focuses on data-driven SSL losses.

\textbf{Contrastive learning.} Constrastive learning~\cite{chen2020simple,he2020momentum,henaff2020data,hjelm2018learning,shen2022connect,tian2020contrastive} (CL) typically transforms an input image and asks the embeddings of the input image and its transformation (the \textit{positive} pair) to be close to one another and far apart from embeddings of other images (\textit{negatives}) via a noise contrastive estimation~\cite{gutmann2010noise,oord2018representation} (NCE) loss. While performant on image-level recognition tasks~\cite{chen2020big,khosla2020supervised}, CL requires non-trivial modification to extend to pixel-level segmentation tasks, as described below.

\textbf{Negative-free representation learning.} In several applications, \textit{true} negative samples may be difficult to construct~\cite{huynh2022boosting}. For example, when learning on intra-domain internal image patches~\cite{dey2022contrareg,park2020contrastive}, non-local spatial positions may be semantically similar, but NCE objectives push their embeddings apart, leading to false negative pairs introducing label noise in the training objective. This drawback may be mitigated via data-driven SSL methods which only use positive samples and avoid \revise{low-diversity (or \textit{collapsed}~\cite{jing2022understanding}) embeddings} solutions via predictor networks and custom backpropogation~\cite{Chen_2021_CVPR,grill2020bootstrap} and careful regularization~\cite{bardes2021vicreg,pmlr-v139-zbontar21a}. However, as above, these methods operate on \textit{global} image embeddings and require modification for pixel-level tasks.

\textbf{Spatial self-supervision.} Towards downstream segmentation, recent work~\cite{Alonso_2021_ICCV,Hu_2021_ICCV,lee2021supervised,liu2022reco,Wang_2021_ICCV,Zhao_2021_ICCV,Zhong_2021_ICCV} encourages local single-scale features either within an image or across images to cluster semantically by constructing positive pairs using ground-truth labels.
To incorporate \textit{local} and \textit{multi-scale} spatial considerations into unpaired image translation and registration~\cite{dey2022contrareg,park2020contrastive} imposed contrastive losses on randomly-sampled layer-wise encoder features by considering corresponding spatial indices as positives and all other locations as negatives. Our work builds on this by instead only considering temporal positives (described below) in the layerwise losses alongside custom regularization which avoids \revise{low-diversity} decoder embeddings observed with naive application in Fig.~\ref{fig:simmap}.

\looseness=-1
\textbf{Temporal self-supervision.} SSL methods developed for video achieve high performance by exploiting temporally consistent transformation~\cite{bai2020can} and temporal pretext task~\cite{cho2020self,Jenni_2021_ICCV,wang2020self}. However, longitudinal biomedical image time-series have sparser sampling (typically 2--5 timepoints/subject) and have greater spatial extents (volumes instead of images), which leads to distinct modeling considerations. 

Emerging biomedical methods~\cite{couronne2021longitudinal,ouyang2021self,zhao2021longitudinal} enforce smooth trajectories for subject-wise images in the encoder latent space and deploy their methods on image-level downstream tasks such as disease classification and age regression. 
However, these methods focus on learning a global embedding, without a clear extension to pixel-level tasks such as segmentation.

\looseness=-1
\textbf{Biomedical image segmentation.} Major challenges specific to biomedical segmentation include: (1) large 3D volumes; (2) limited sample sizes and annotations; and (3) non-i.i.d. longitudinal acquisitions tracking temporal anatomical changes. To these ends, conventional approaches use a combination of intensity-based probabilistic models and registration-driven atlas-based models~\cite{aljabar2009multi,iglesias2015multi,lotjonen2010fast,wang2012multi}. In particular, longitudinal image analysis typically makes use of one or few longitudinal atlases~\cite{iglesias2016bayesian,kim2013adaptive,reuter2012within,shi2010neonatal,shi2011infant}, which motivates the one and few-shot segmentation settings benchmarked in this paper, respectively.

\looseness=-1
More recently, deep segmentation networks achieve strong performance~\cite{billot2021synthseg,billot2022robust,milletari2016v,myronenko20183d,ronneberger2015u} given enough training volumes.
In the low-annotation setting, weakly supervised methods develop custom loss functions~\cite{kervadec2019constrained, li2021point}, but may have drawbracks analogous to the handcrafted SSL losses described above. Fortunately, recent data-driven self and semi-supervised methods are well-suited to pixel-level prediction. 
For example, to pretrain an encoder for segmentation, \cite{chaitanya2020contrastive,zeng2021positional} develop application-specific positive and negative sampling strategies for contrastive training, where 2D slices from similar locations in registered 3D volumes across subjects constitute positive pairs. While these methods have been successful in their applications, they are inherently slice-based methods and are outperformed by well-tuned randomly-initialized 3D baselines on our datasets (Tab. \ref{tab:result_moremetrics}).
Further, these self-supervised biomedical segmentation methods do not explicitly account for non i.i.d acquisitions. Lastly, to our knowledge, existing longitudinal deep learning work developed for biomedical segmentation is currently very specific to its target application\revise{~\cite{gao2018fully, li2021longitudinal, to2021self} (for example, in tasks such as MS lesion change detection~\cite{to2021self}) or require supervised pre-training on annotated cross-sectional datasets~\cite{wei2021consistent}}, whereas we develop a generic self-supervised spatiotemporal representation learning framework for non-i.i.d. longitudinal data which can be applied to any downstream task in principle.

\section{Methodology}
\label{sec:method}
\begin{figure}
    \centering
    \includegraphics[width=\textwidth]{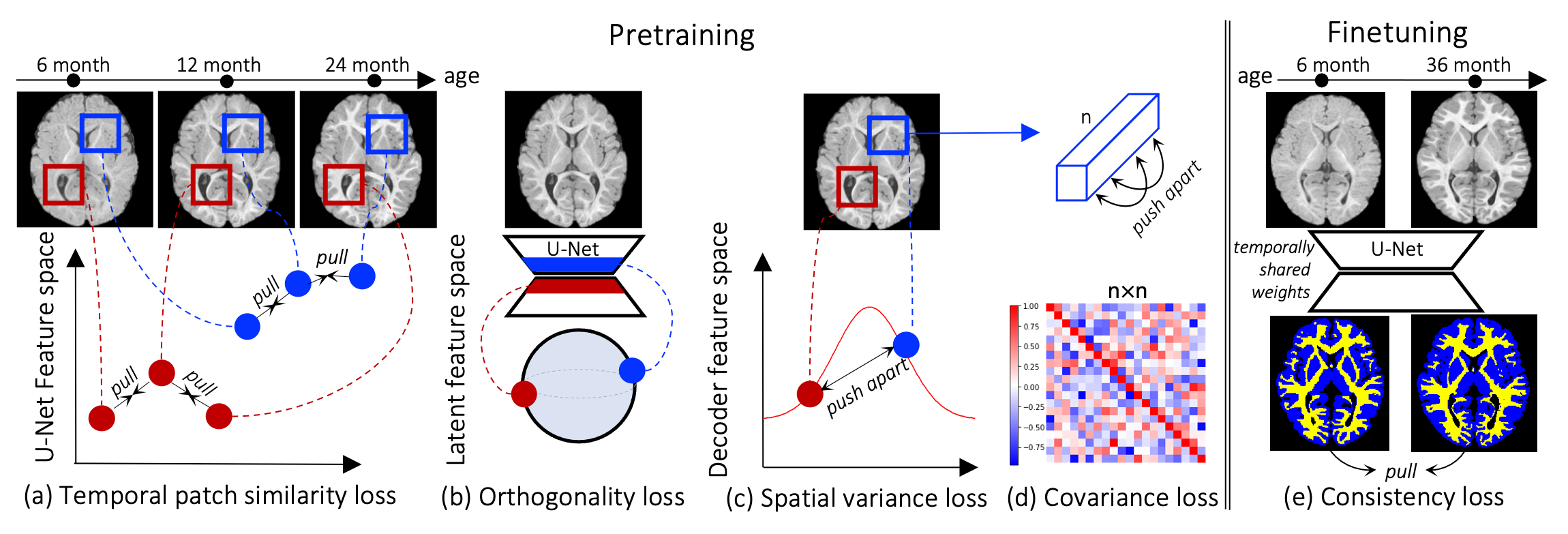}
    \caption{\textbf{Overview of proposed self-supervision.} Given nonlinearly-registered temporal images of a subject, \textbf{(a)} we assume that corresponding spatial locations in various network layers should have similar representations. As U-Net skip connections can cause \revise{degenerate decoder embeddings (see App~\ref{app:needforreg})}, we \textbf{(b)} encourage the decoder bottleneck to be orthogonal to encoder bottleneck and regularize the concatenated decoder features to have \textbf{(c)} high spatial variance and be \textbf{(d)} uncorrelated channel-wise. During fine-tuning, we \textbf{(e)} encourage temporal intra-subject network output consistency.}
    \label{fig:overview}
\end{figure}

Fig.~\ref{fig:overview} \revise{illustrates} the proposed framework. The base U-Net architecture is pretrained end-to-end in a self-supervised manner with intra-subject \revise{spatiotemporal} losses. 
On convergence, the pretrained network parameters serve as an initialization for training downstream local spatiotemporal pixel-level tasks (e.g., registration or segmentation). This work will focus on downstream segmentation in the one, few, and full-shot regimes.
The main similarity loss is applied on multiscale local patches from different timepoints of the same subject, which attracts features in corresponding locations in separate timepoints together. The high-dimensional U-Net features corresponding to the boxes \revise{at the same locations} in Fig.~\ref{fig:overview}a should maintain high similarity despite varying appearance. The feature regularizers (Fig.~\ref{fig:overview}b,c,d) avoid \revise{degenerate} U-Net decoder embeddings in the patch similarity training and the output regularizer (Fig.~\ref{fig:overview}e) encourages finetuning consistency on unannotated data. 

\textbf{Setup.} The unlabeled dataset is a collection of $N$ subjects, where each subject has at least two longitudinal image acquisitions available during pretraining. 
During every iteration, a pair of images $(\mathcal{X}_{j}^{i}, \mathcal{X}_{j+1}^{i})$ are randomly sampled from subject ${i}\in\{1,2,\dots,N\}$ at distinct timepoints ${j}$ and ${j+1}$, where $j\in\{1,2,\dots,T_i-1\}$. $T_i$ indicates the number of registered images from subject $i$, and $\mathcal{X} \in \mathcal{R}^{W\times H\times D \times C}$ is a 3D volume of spatial dimension $W\times H\times D $ and $C$ channels. These channels are typically multi-modality acquisitions (e.g., T1w and T2w MRI from the same subject).  

\noindent\textbf{Spatiotemporal patchwise similarity loss.}
We aim to associate the local embeddings of $\mathcal{X}_{j}^{i}$ and $\mathcal{X}_{j+1}^{i}$ such that the representations are longitudinally-aware.
To this end, a weight-sharing 3D U-Net ${G}$ takes both images as input and produce a set of multi-scale CNN features $\{v\}_L$, where each element $v_{lij}=G^l(x_{j}^{i}) \in \mathcal{R}^{W_l\times H_l \times D_l\times C_l}$ indicates the output of the $l$th layer of interest. 
${M}$ feature vectors are then randomly sampled from the 3D spatial indices of the feature map, where each feature vector ${v_{lij}^m}$ represents a local patch of the input image\revise{. These} patch-wise activations from multiple layers of $G$ form hierarchical representations of local regions of the input image.

To maximize the similarity of corresponding local features (e.g., blue boxes in Fig.~\ref{fig:overview}\revise{a}) without negative samples, we use a projector MLP ${f}$ and a predictor head ${p}$ and extend~\cite{Chen_2021_CVPR} to patchwise operation such that matching spatial indices have high agreement.
The patchwise similarity loss between two representations is defined as follows:
$
    \mathcal{L}(v_{lij}^m, v_{lik}^m) = \frac{1}{2}\mathcal{D}(p_1,z_2)+\frac{1}{2}\mathcal{D}(p_2,z_1),
$
 where
$\mathcal{D}(p,z) = - \frac{p}{||p||_2}\cdot \frac{z}{||z||_2}$, 
$z_1, z_2 = f(v_{lij}^m), f(v_{lik}^m)$ and 
$p_1, p_2 = p(z1), p(z2)$.
The total loss is an average of all sampled patches, across multilayer features:
\begin{align}
\mathcal{L}_{sim} =\frac{1}{L} \sum_{l\in \{1,2,\dots L\}} \frac{1}{M} \sum_{m\in \{1,2,\dots M\}} \mathcal{L}(v_{lij}^m, v_{lik}^m).
\end{align}

\textbf{Architectural challenges in multi-scale representation learning}. $\mathcal{L}_{sim}$ applied to the hidden layers of a U-Net is found to maximize patchwise similarity of the encoder features (as expected) but lead to \revise{low-diversity and semantically-incoherent representations} in the decoder layers as observed from the similarity maps in Fig.~\ref{fig:simmap}B \revise{and empirical observations in App.~\ref{app:needforreg}.} We \textit{speculate} that this is \revise{partially attributable} to U-Net skip connections, which lead to a degenerate solution where the encoder learns representations which are good enough to minimize the decoder losses and the decoder layers do not have to learn \revise{useful} representations \revise{which transfer}. To this end, we develop several regularization strategies such that the decoder layers \revise{obtain diverse and semantically-coherent embeddings}.

\textbf{Orthogonality.} During model prototyping, we empirically observe that \revise{low-diversity embeddings} first originate in the \revise{U-Net bottleneck (Fig.~\ref{fig:simmap}B cols 4,5 and App. Fig.~\ref{fig:collapse_featuremaps_and_svd} rows 2,3) which are then upsampled hierarchically through the decoder}. We therefore encourage decoupled \revise{bottleneck} features between the encoder and decoder. Revisiting the U-Net skip-connection, the encoder features from the $l$-th layer $v_e\in \mathcal{R}^{W_l\times H_l \times D_l\times C_l}$ are concatenated with the upsampled decoder features
$v_{d}\in \mathcal{R}^{W_l\times H_l \times D_l\times C_d}$, followed by a convolution to attain the same feature dimension as $v_e$. This yields
$v_{\hat{d}}=\texttt{Conv}(\texttt{Concat}(v_e, v_{d})) \in \mathcal{R}^{W_l\times H_l \times D_l\times C_l}$ and by regularizing for orthogonality between $v_e$ and $v_{\hat{d}}$ in the projector embedding space, we implicitly encourage $v_d$ to learn better representations instead of converging to degenerate solutions.
The orthogonality loss is defined as 
\begin{align}
    \mathcal{L}_O(z_e, z_{\hat{d}}) = \frac{1}{M}\sum_{m\in\{1,\dots M\}}\frac{z_e^m}{||z_e^m||_2}\cdot \frac{z_{\hat{d}}^m}{||z_{\hat{d}}^m||_2},
\end{align} where $z_e^m$ and $z_{\hat{d}}^m$ are projected representations via $f$ at sampling location $m$, from the matched encoder/decoder layers, respectively. 

\textbf{Variance and covariance.} We further encourage spatial variation and channel-wise decorrelation of local decoder features to \revise{avoid degenerate representations.} 
We extend the variance and covariance terms of~\cite{bardes2021vicreg} towards patchwise multi-\revise{layer} operation and apply them to the decoder layers. A standard deviation loss encourages spatial feature variation above a threshold of $\eta=1$ and is defined as 
\begin{align}
    \mathcal{L}_{S}(z) = \frac{1}{k}\sum_{l\in\{L-k,\dots L\}} (\max(0, \eta - S(z_l)),
\end{align} where $S(z_l)=\sqrt{Var(z_{l})+\epsilon}$ is the standard deviation of $M$ randomly-sampled projected features from the $l$-th layer and $\epsilon=10^{-4}$ is added for numerical stability. A covariance regularization decorrelates channelwise activations in the decoder to prevent \revise{low-diversity} embeddings by minimizing 
\begin{align}
    \mathcal{L}_C(z)=\frac{1}{k}\sum_{l\in\{L-k,\dots L\}}\frac{1}{n}\sum_{u\neq v}[C(z)]_{u,v}^2,
\end{align}
where $C(z) = \frac{1}{n-1} \sum_{i}^n (z_i-\overline{z})(z_i-\overline{z})^T$ is the covariance matrix of $n$-D representation $z$ and $(u,v)$ indicates its off-diagonal indices.

\noindent\textbf{Reconstruction.} To further pretrain the decoder, we investigate reconstruction losses commonly used in unsupervised learning.
However, as high-resolution information is passed through skip connections, a reconstruction loss for a U-Net is near-trivially minimized and does not address \revise{degenerate solutions}. Therefore, inspired by denoising autoencoders~\cite{vincent2008extracting}, we encourage network equivariance and invariance to geometric ($\mathcal{A}_g$) and intensity-based transformations ($\mathcal{A}_i$), respectively, of the input image $x$, by modifying the reconstruction objective to a denoising loss as,
\begin{align}
\mathcal{L}_{rec}=\|G(\mathcal{A}_i(\mathcal{A}_{g}(x)))-\mathcal{A}_g(x)\|_2^2.
\end{align}

\noindent\textbf{Finetuning with longitudinal consistency regularization.}
In longitudinal segmentation, if the finetuning data do not cover the overall age range, the network may perform well on the finetuned timepoints, but may perform poorly on unseen ages even when pretrained on unannotated data across all ages. 
We therefore develop a self-supervised longitudinal consistency regularization term applied at the network output during finetuning to increase the intra-subject agreement. Given registered and unannotated image volumes, we formulate a segmentation prediction consistency loss as, 
\begin{align}
\mathcal{L}_{cs} = 1 - \texttt{Dice}(G(x_j^i), G(x_{j+1}^i)),
\end{align}
which is minimized alongside the supervised segmentation term below during finetuning.

\noindent\textbf{Total objective.}
During pretraining, our overall objective function is a weighted sum of the patch similarity loss, reconstruction loss, and three regularizations, and is defined as,
\begin{align}
\mathcal{L}_{PT} = \lambda \mathcal{L}_{sim} + \alpha \mathcal{L}_{rec} + \mu \mathcal{L}_S(z)+ \gamma \mathcal{L}_C(z) + \beta \mathcal{L}_O(z).
\end{align}
During finetuning, we use a combined dice and cross entropy loss on supervised training pairs as, 
\begin{align}
    \mathcal{L}_{sup}=(1 - \texttt{Dice}(G(x), y)) + \texttt{CE}(G(x), y),
\end{align} where $y$ is the groundtruth segmentation label. Lastly, we additionally use the segmentation consistency loss $\mathcal{L}_{cs}$ for non i.i.d inputs, which forms our final finetuning loss $\mathcal{L}_{FT}=\mathcal{L}_{sup}+\mathcal{L}_{cs}$.

\begin{figure}[t]
    \centering
    \includegraphics[width=\textwidth]{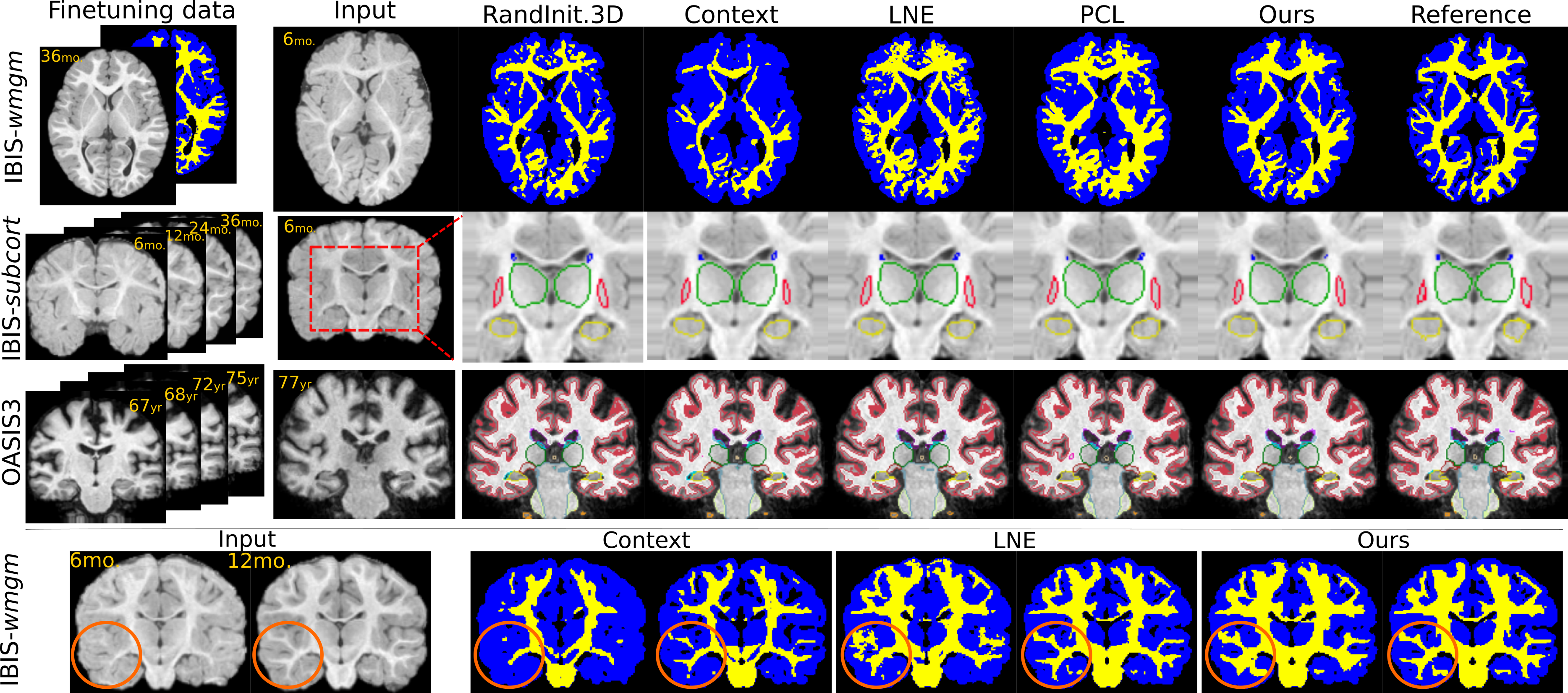}
    \caption{\textbf{One-shot segmentation}. \textbf{Top 3 rows:} Once pretrained on all unlabeled data, all benchmarked methods are finetuned on either a single annotated image (IBIS-\textit{wmgm}) or a single annotated subject (IBIS-\textit{subcort} and OASIS3). When deployed on other subjects at different ages, our method yields improved segmentation performance. \textbf{Bottom row:} When finetuned only on a single 36 month-old image, our method generalizes to unseen timepoints by leveraging temporal consistency.  
    }
    \label{fig:seg_example}
\end{figure}

\section{Experiments}
\label{sec:experiments}
\looseness=-1
\textbf{Data and segmentation tasks.} We conduct experiments on two de-identified longitudinal neuroimaging datasets, and specifically design three tasks to benchmark different extents of biomedical domain gaps between the finetuning and testing data. \revise{The main body of this work focuses on one-shot segmentation using one annotated subject (a common medical image analysis setting). Benchmarks on few-shot and fully-supervised segmentation tasks are provided in Appendix \ref{app:more_results}.} For both datasets, we perform a train/validation/test split on a subject-wise basis with 70\%, 10\% and 20\% of the participants. The validation set is used for model and hyperparameter selection and results are reported on a held-out test set.
ANTs~\cite{avants2008symmetric,avants2010optimal} is used to perform inter-subject affine alignment for all experiments, followed by intra-subject deformable registration to obtain accurate spatiotemporal correspondence for $\mathcal{L}_{sim}$ and $\mathcal{L}_{cs}$ calculation. All images are skull-stripped, bias-field corrected, and intensity-normalized. Further data preprocessing and splitting details are described in Appendix~\ref{app:dataprep}.

\looseness=-1
\underline{OASIS3}~\cite{lamontagne2019oasis} is a publicly-available dataset consisting of 1639 \revise{brain} MRI scans of 992 longitudinally imaged subjects\revise{. Each subject has} 1--5 temporal acquisitions over a $\sim5$-year long observation window, resulting in an aging cohort over the span 42 to 95 years \revise{which includes} cognitively normal and mildly impaired individuals alongside subjects with Alzheimer's Disease. \revise{On OASIS3, we tackle whole-brain segmentation using the FreeSurfer label convention~\cite{fischl2002whole}.} \textit{Cross-sectional} FreeSurfer anatomical segmentation was done as part of the data release. Observing strong temporal inconsistency (see \revise{App.~\ref{app:more_preproc}}), we further perform \textit{longitudinal} FreeSurfer~\cite{reuter2012within} to improve the temporal consistency of the reference segmentation. 
We exclude labels that have less than 100 voxels in all subjects, which results in 33 labels for segmentation training and evaluation. Finetuning \revise{for one-shot segmentation} is performed on a single FreeSurfer-annotated subject with four timepoints.

\underline{IBIS} is an infant brain imaging study, which longitudinally acquires 1272 structural T1w/T2w MRI from 552 infants across both controls and infants at a high-risk for Autism Spectrum Disorder (ASD) over a span of 3 to 36 months of age. We tackle two distinct tasks:  subcortical segmentation (\textit{IBIS-subcort}) and white/gray matter tissue segmentation (\textit{IBIS-wmgm}). For \noindent\textit{IBIS-subcort}, a multi-atlas method~\cite{swanson2017subcortical} cross-sectionally segments sub-cortical grey matter (relevant to ASD~\cite{shen2022connect}) into 13 structures of interest, which are then followed by manual corrections. \revise{For one-shot benchmarking}, finetuning is performed only on a single longitudinally-labeled subject, similar to OASIS3 above. Further, we use the \textit{IBIS-wmgm} setting to simulate a real-world use-case detailed in App.~\ref{app:wmgm_motivation}. Briefly, brain MRI segmentation into grey/white matter is straightforward at 24-36 months of age due to the presence of anatomical edges in the images. \revise{However, $\sim$}6-month-old grey/white matter brain segmentation remains elusive without manually labeled datasets for supervision due to white matter myelination leading to isointense appearance at that age~\cite{sun2021multi} 
(e.g., Fig.~\ref{fig:overview}a and \revise{App.~\ref{app:wmgm_motivation}}). We therefore investigate finetuning all benchmarked methods on a single 36 month old image which can be reliably segmented with~\cite{puonti2016fast} and then evaluate segmentation deployment with a strong domain shift on 6 month old isointense images and labels (whose ground truth labels are generated by a fully supervised external model~\cite{zeng2018multi}).

\textbf{Baselines and Evaluation Strategies.} We analyze segmentation performance and longitudinal consistency against well-tuned randomly initialized 2D/3D U-Nets (\texttt{RandInitUnet.2D/3D}) and various high-performing self-supervised pretraining methods. \revise{These include} the pretext-task based \texttt{Context Restoration}~\cite{chen2019self}, longitudinal representation learning based \texttt{LNE}~\cite{ouyang2021self}, along with the contrastive learning based \texttt{GLCL}~\cite{chaitanya2020contrastive} and  \texttt{PCL}~\cite{zeng2021positional} methods which operate on image slices. We also repurpose \texttt{PatchNCE}~\cite{park2020contrastive} \revise{for segmentation} to evaluate its generic representation learning capabilities. All methods are pretrained and finetuned with both geometric and intensity-based augmentation, and share the same network architecture. 

\looseness=-1
We quantify network performance via commonly used scores such as the Dice coefficient, IoU, and the 95-th percentile of the Hausdorff distance. More importantly, we also quantify the longitudinal agreement between intra-subject non-linearly registered temporal segmentations via scores such as the spatiotemporal consistency of segmentation~\cite{li2021longitudinal} $STCS= \frac{2|S_1\cap S_2|}{|S_1|+|S_2|}$ where $S_1$ and $S_2$ are temporal segmentation predictions from non-linearly registered input images; and the absolute symmetrized percent change~\cite{reuter2012within} $ASPC = 100\frac{|V_2-V_1|}{0.5(V_1+V_2)}$, where $V_1$ and $V_2$ are the volume of a structure calculated from $S_1$ and $S_2$. We also report $STCS$ and $ASPC$ on the groundtruth segmentations as a reference. 

\noindent\textbf{Implementation details.} We train a 3D U-Net~\cite{ronneberger2015u} as the base image-to-image architecture with four levels of up/down sampling and repeated Conv-BN-ReLU blocks (\revise{all} architectural details are provided in App.~\ref{app:addn_impl_dets}). The projector head consists of a 3-layer MLP with 2048 nodes per layer. Following~\cite{Chen_2021_CVPR}, we apply batch normalization after each MLP, followed by ReLU activation, and $l_2$ normalization of the final activation. The predictor is a 3-layer bottlenecked MLP with widths of 2048-256-2048. 
We apply geometric (left-right flip, random affine warps) and intensity augmentations including random blur, noise, gamma contrast enhancement. Additional MRI-specific augmentations with random bias field and motion artifacts are also applied and are followed by $128^3$ random spatial cropping.
We use a batch size of 3 crops \revise{and} an initial learning rate of $2 \times 10^{-4}$ \revise{for both pretraining and finetuning}. 
All networks are trained with the Adam optimizer ($\beta_1=0.9$ during pretraining and $\beta_1=0.5$ during finetuning and $\beta_2=0.999$ in both settings) on a single Nvidia RTX8000 GPU (45GB vRAM). The networks are pretrained for a maximum of $30,000$ steps and the best model based on validation performance is used for fine-tuning for another $35,000$ steps, alongside linear learning rate decay. All experiments are run on a fixed random seed due to limited computational budgets. Based on the ablation analysis in Tab.~\ref{tab:ablation}, we empirically choose $\lambda=1,\alpha=10,\gamma=1e{-3}, \beta=100$ for all datasets, and use $\mu=10^{-2}$ for OASIS3, $\mu=10^{-3}$ for IBIS. 
Further details on the configurations, implementation of our method and other baselines are provided in \revise{Appendix~\ref{app:addn_impl_dets}}.

\setlength{\tabcolsep}{4pt}
\begin{table}[t]
\begin{minipage}{\linewidth}
\footnotesize
\centering
\caption{\textbf{One-shot segmentation benchmarking} of  performance (median IoU \& HD95) and longitudinal consistency (ASPC). Further scores alongside means and std. dev. \revise{are provided in Supplemental Table \ref{tab:one_shot_mean_std}. \textbf{Few-shot} and \textbf{fully-supervised} results are provided in Suppl. Tabs. \ref{tab:finetune_0.1} and \ref{tab:finetune_1.0}, respectively.}}
\begin{tabular}{lccccccccc}
\toprule
\multicolumn{1}{c}{\multirow{2}{*}{Method}} & \multicolumn{3}{c}{IBIS-subcort}   & \multicolumn{3}{c}{IBIS-wmgm}    & \multicolumn{3}{c}{OASIS3}         \\
  &   IoU $\uparrow$               & HD95  $\downarrow$               & ASPC$\downarrow$ & IoU$\uparrow$                   & HD95$\downarrow$                &ASPC$\downarrow$ & IoU $\uparrow$                   & HD95  $\downarrow$              &  ASPC$\downarrow$  \\ \midrule
GT   & -         & -       & 7.819    & -      &  -     & -       & -      &-           & 3.947 \\
RandInitUnet.2D     & 0.707    & 4.485    & 7.127    & 0.510  & 3.274  & 11.506  & 0.687  & 2.545      & 10.189 \\
RandInitUnet.3D     & 0.720    & 2.892    & 4.644    & 0.560  & 3.788  & 5.515   & 0.715  & 2.206      & 2.789 \\
Context Restore~\cite{chen2019self}     & 0.711    & 4.403    & 7.831    & 0.444  & 8.273  & 29.235  & 0.717  & 3.323      & 5.577 \\
LNE~\cite{ouyang2021self}         & 0.736    & 3.033    & 5.866    & 0.563  & 3.201  & 5.352   & 0.726  & 1.988      & 8.836 \\
GLCL~\cite{chaitanya2020contrastive}         & 0.718    & 3.203    & 5.514    & 0.550  & 4.112   & 8.472  & 0.695 & 2.264      & 4.622 \\
PCL~\cite{zeng2021positional}         & 0.713    & 3.270   & 5.610     & 0.562  & 4.974  & 10.648  & 0.707  & 2.327     & 4.850  \\
PatchNCE~\cite{park2020contrastive}    & 0.743   & 1.266   & 5.780     & 0.607  & 4.344  & 3.782  & 0.738  & 2.275      & 3.114  \\
Ours w/o $\mathcal{L}_{cs}$    & 0.754   & \textbf{1.145}      & 5.483  & 0.614  & 3.291  &\textbf{ 2.462}  &\textbf{ 0.739 }     & \textbf{1.940} & \textbf{2.729} \\
Ours w/ $\mathcal{L}_{cs}$  &\textbf{ 0.757 }  & 1.178     & \textbf{4.475} & \textbf{0.676} & \textbf{3.237} & 4.155 & 0.737 & 2.094 & 2.754 \\    
\bottomrule  
\end{tabular}
\label{tab:result_moremetrics}
\end{minipage}
\begin{minipage}{\linewidth}
    \centering
    \includegraphics[width=\textwidth]{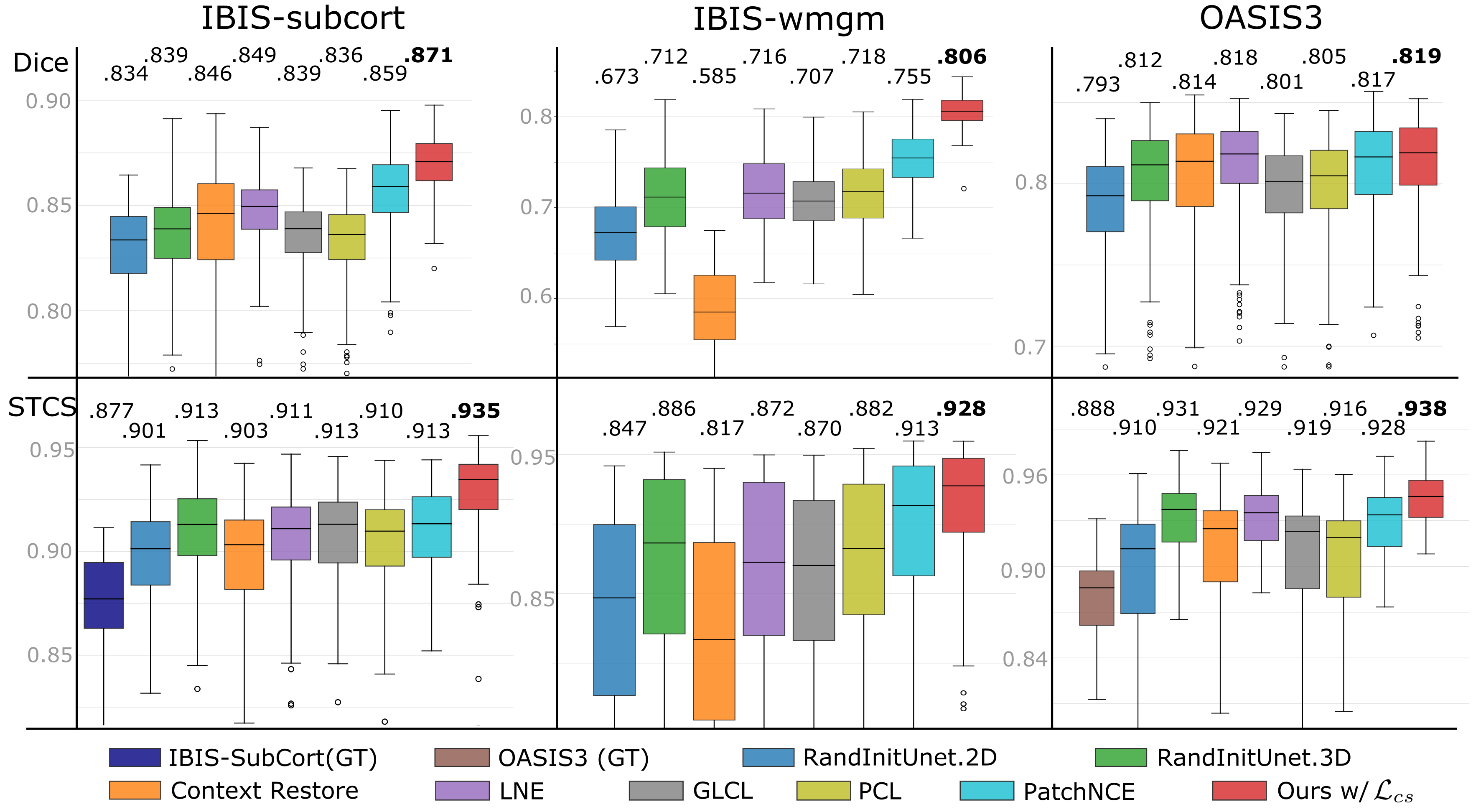}
	\captionof{figure}{\textbf{One-shot segmentation benchmarking} quantifying performance with the Dice coefficient (\textbf{top}) and the spatiotemporal consistency of segmentation (\textbf{bottom}), visualizing the means and standard deviations alongside median values overlaid on the top of each subfigure (higher is better). \revise{\textbf{Few-shot} and \textbf{fully-supervised} results are provided in Suppl. Tabs. \ref{tab:finetune_0.1} and \ref{tab:finetune_1.0}, respectively.}}
    \label{fig:result_boxplot}
\end{minipage}
\end{table}

\looseness=-1
\textbf{Segmentation and longitudinal consistency results.}
Fig.~\ref{fig:seg_example} qualitatively demonstrates improved generalization using our method on unseen longitudinal data (row 1-3), especially on data displaying rapid intra-subject temporal developments (IBIS-\textit{{wmgm,subcort}}). 
These improvements are consistent with the quantitative results presented in Fig.~\ref{fig:result_boxplot}/Tab.~\ref{tab:result_moremetrics} which indicate both improved segmentation performance and longitudinal consistency. In the strong domain shift setting of \textit{IBIS-wmgm}, we see a near ten-point increase in median dice over most baselines. With moderate shifts in \textit{IBIS-subcort}, we see appreciable increases in performance and consistency. We note that brain segmentation on adult brain MRI data from OASIS3 is a comparatively easier task as adult neuroimages do not significantly change appearance between imaging sessions. Therefore, several baselines are able to match (but not exceed) the segmentation performance of our method on OASIS3. However, all baselines are outperformed by ours on all datasets in terms of longitudinal-consistency which is essential to non-i.i.d. statistical analysis.
In particular, both our pretraining (\texttt{Ours w/o} $\mathcal{L}_{cs}$) and finetuning (\texttt{Ours w/}  $\mathcal{L}_{cs}$) methods show STCS and ASPC improvements over all of the compared settings. Fig.~\ref{fig:seg_example} (bottom row) shows an example of temporal predictions on two unseen timepoints from IBIS-wmgm\revise{. T}he predictions from Context Restoration~\cite{chen2019self} match only the input image intensity and lack anatomical and longitudinal consistency, \revise{LNE~\cite{ouyang2021self} introduces} false positive predictions in temporal lobe (within the orange circle), and \revise{the proposed} method yields a more spatiotemporally and anatomically consistent segmentation. \revise{Beyond one-shot segmentation, we also observe gains in the \textbf{few-shot} and \textbf{fully-supervised} segmentation settings in Appendix \ref{app:more_results} and Suppl. Tabs. \ref{tab:finetune_0.1} and \ref{tab:finetune_1.0}, respectively.} 

\noindent\textbf{Qualitative self-supervised spatiotemporal similarity.}
In Fig.~\ref{fig:simmap}, we qualitatively examine the learned visual representations of the proposed method via intra-subject temporal self-similarity (\textbf{C}) and compare it to two of its variants which either use contrastive learning with unsupervised negatives (A) or negative-free representation learning (B). 
We calculate the per-layer multiscale feature self-similarity between
the {\color{red} query} and each {\color{blue}key} from the intra-subject feature maps at a different age (blue box). In \underline{row A}, we see that assuming that all spatial indices not in correspondence constitute negative pairs leads to highly-positionally dependent representations in the decoder which carry low semantic meaning (e.g., in the adult data, the similarity to localize to the ventricles in the coronal view). By discarding all negatives in \underline{row B}, we observe \revise{semantically-incoherent and low-diversity embeddings} and artifacts in the decoder \revise{layers} on both datasets. Finally, with careful regularization in row C, our methods discards all negative pairs and attains semantically and positionally relevant representations.

\looseness=-1
\noindent\textbf{Ablations.} As the proposed method consists of several moving parts, an ablation analysis is conducted over different model configurations, hyperparameters, and loss functions, reported in Tab.~\ref{tab:ablation} consisting of average dice coefficients. The combination of all proposed components yields optimal results. Further ablations and baseline tuning results are reported in Appendix~\ref{app:more_results}.

\underline{Row A} starts with a base setting where only four encoder layers from the U-Net are selected for $\mathcal{L}_{sim}$ computation and a small MLP width of 256 is used for the projection and prediction heads. 
Here, IBIS-subcort and OASIS3 results are competitive with randomly initialized U-Net, as expected given the lack of auxiliary losses, data augmentation, and regularizations. However, the IBIS-wmgm experiment already shows a 2\% improvement over random initialization, indicating benefits of using patchwise similarity losses for better out-of-distribution generalization even with suboptimal setups. 

\underline{Row B:} With larger projector and predictor networks, we observe improvements on two out of three datasets, which is consistent with trends observed on natural images~\cite{chen2020big,Chen_2021_CVPR}. 

\underline{Rows C--F:} On adding decoder layers to $\mathcal{L}_{sim}$ and introducing $\mathcal{L}_{rec}$ alongside data augmentation (without any regularization), we typically observe inconsistent dataset-specific trends which arise from \revise{unregularized} representations (e.g., Fig.~\ref{fig:simmap}B). We speculate that \revise{a poorly trained decoder} (due to a lack of regularization) may be equivalent to random initialization \revise{in the context} of pretraining for segmentation tasks. However, a combination of these components (\underline{Row F}) leads to an appreciable increase in performance. 

\looseness=-1
\underline{Rows G--K:} When orthogonal regularization and/or covariance/variance regularization is used, we observe the best performance when they are applied together alongside augmentation and $\mathcal{L}_{Rec}$. In rows J and K, we observe that different hyperparameters are optimal for OASIS3 and IBIS (which already outperform all baseline methods in Tab.~\ref{tab:result_moremetrics}), which is intuitive as these are drastically different cohorts. 

\underline{Row L:} Finally, the overall proposed model is achieved when $L_{cs}$ is added to the finetuning objective, which yields strong improvements for IBIS-\{\textit{wmgm,subcort}\} and maintains OASIS3 performance.

\setlength{\tabcolsep}{3.5pt}
\begin{table}[t]
\footnotesize
\centering
\caption{\textbf{Ablation analysis} of our method over loss layers, projection+prediction layer widths (\#\texttt{MLP}), loss functions ($\mathcal{L}_{Rec}$, $\mathcal{L}_{cs}$), use of augmentation, and hyperparameters ($\beta, \mu, \gamma$). Mean dice is used for quantification on all datasets. *$\mu=10^{-3}$ on IBIS-\{\textit{wmgm, subcort}\} and $\mu=10^{-2}$ on OASIS3.}
\begin{tabular}{cccccccccccc}
\toprule
Exp & Loss Layers    & \#\texttt{MLP}   & $\mathcal{L}_{Rec}$       & Aug.       & $\beta$       &$\mu$     & $\gamma$  & $\mathcal{L}_{cs}$  & IBIS-\textit{subcort} & IBIS-\textit{wmgm}    & OASIS3      \\ \midrule
A & Enc      & 256   &              &              &            &           &           &      & 0.829(0.068)          & 0.733(0.062)          & 0.783(0.16)          \\
B & Enc      & 2048  &              &              &            &           &           &      & 0.849(0.060)          & 0.732(0.073)          & 0.809(0.13)          \\
C & Enc      & 2048  & $\checkmark$ &              &            &           &           &      & 0.859(0.058)          & 0.713(0.079)          & 0.811(0.13)          \\
D & EncDec     & 2048  & $\checkmark$ &              &            &           &            &    & 0.860(0.058)          & 0.718(0.066)         & 0.810(0.13)          \\
E & EncDec      & 2048  &               & $\checkmark$  &         &            &          &    & 0.858(0.060)          & 0.724(0.077)          & 0.809(0.13)           \\
F & EncDec     & 2048  & $\checkmark$ & $\checkmark$ &            &           &           &    & 0.856(0.060)          & 0.739(0.067)          & 0.812(0.13)          \\
G & EncDec     & 2048  & $\checkmark$ &              & 100        &           &           &    & 0.857(0.062)          & 0.728(0.074)          & 0.809(0.13)          \\
H & EncDec     & 2048  & $\checkmark$ &              &            & $10^{-3}$ & $10^{-3}$ &    & 0.845(0.063)          & 0.739(0.074)          & 0.804(0.13)          \\
I & EncDec     & 2048  & $\checkmark$ &              & 100        & $10^{-3}$ & $10^{-3}$ &    & 0.859(0.056)          & 0.735(0.061)          & 0.811(0.13)          \\
J & EncDec     & 2048  & $\checkmark$ & $\checkmark$ & 100        & $10^{-3}$ & $10^{-3}$ &    &{0.863(0.057)} & 0.758(0.062) & 0.808(0.14)          \\
K & EncDec     & 2048  & $\checkmark$ & $\checkmark$ & 100        & $10^{-2}$ & $10^{-3}$ &    & 0.853(0.055)          & 0.745(0.058)          & \textbf{0.813(0.13)}\\
L & EncDec     & 2048  & $\checkmark$ & $\checkmark$ & 100        &   *        & $10^{-3}$ & $\checkmark$   & \textbf{0.870(0.052)}       &   \textbf{0.806(0.030)}   & \ 0.810(0.13) \\
\bottomrule
\end{tabular}
\label{tab:ablation}
\end{table}

\section{Discussion}
\label{sec:discussion}
\noindent\textbf{Limitations and future work.} The presented work opens up many follow-up questions which will be tackled in future work: (1) Our proposed losses enable better training and performance, but require the tuning of several regularization weights and layer selections which may reveal dataset specific patterns (e.g., rows J, K in Tab. \ref{tab:ablation}). The weights and layers selected here were chosen based on limited exploratory experiments on the validation sets due to computational budgets and future work will exhaustively search the hyperparameter space for optimal performance. (2) Our pretraining assumes accurate non-linear intra-subject registration, which may be non-trivial in edge cases like modalities with strong distortion and artifacts (e.g., eddy corruption in diffusion MRI). However, when studying pre and post-operative imaging (e.g., surgical excision of lesions), large topological changes break the assumptions of our model and will require the development of lesion-masked positive patch sampling methods. 
(3) We use a two-stage pre-training and finetuning approach and it is plausible that the proposed method can be reduced to a single-stage combined framework.
(4) While this paper focused on downstream segmentation finetuning, the pretraining framework is generic to any pixel-level task (e.g., registration) and its extension to such tasks will be explored. (5) While the proposed methods yield strong longitudinal segmentation consistency improvements across all datasets, we note that absolute segmentation performance gains in an elderly cohort (OASIS3) are modest in comparison to the rapidly developing infant dataset (IBIS) where higher gains are achieved. Future work will further investigate these performance differences between populations of differing temporal trends. (6) We extended the negative-free framework of~\cite{Chen_2021_CVPR} to patchwise operations for its relative simplicity and it is plausible that other negative-free similarity terms~\cite{bardes2021vicreg,grill2020bootstrap,pmlr-v139-zbontar21a} may further improve results. (7) In our data preparation for pretraining, subject-wise image time-series were registered to a single time-point instead of a subject-specific template, which is known to increase statistical bias~\cite{reuter2012within}. 

\looseness=-1
\textbf{Limited scope.} The proposed method generically applies to medical image time-series and we do not anticipate negative impacts beyond those that currently exist for segmentation methods. However, while our tasks have impacts on understanding real-world disease mechanisms, we cannot claim any further insight into differences between subpopulations, as such analysis requires close collaboration with clinicians, neuroscientists, and biostatisticians, which is beyond the scope of this work.

\looseness=-1
\textbf{Conclusions.} This paper addressed several open questions regarding the self-supervised pretraining and finetuning of image-to-image architectures on \revise{longitudinal} volumes using objective functions which exploit both intra-subject spatial and temporal self-similarity. It developed a local negative sample-free framework that trains multiple multi-scale hidden layers of image-to-image architectures that then enabled improved downstream segmentation performance, all while \revise{achieving semantically-meaningful representations} via careful regularization of the decoder activations. During finetuning, it similarly developed a simple consistency-regularization objective which encourages longitudinal agreement between predictions on \revise{unlabeled} data. When applied to large-scale neurodeveloping and neurodegerative longitudinal \revise{images}, the proposed framework yielded improved segmentation performance and temporal consistency, both of which are crucial to statistical analyses of mechanisms of interest such as Alzheimer's Disease (OASIS3) and Autism Spectrum Disorder (IBIS).

\section*{Acknowledgements}
The authors are grateful to NIH R01-HD055741-12, R01-MH118362-02S1, 1R01MH118362-01, 1R01HD088125-01A1, R01MH122447, R01-HD059854, U54-HD079124, P50-HD103573, R01ES032294, and the NYS Center for Advanced Technology in Telecommunications (CATT). 

Adult longitudinal T1w MRI data were provided by OASIS-3;
NIH P50 AG00561, P30 NS09857781, P01 AG026276, P01 AG003991, R01 AG043434, UL1 TR000448, R01 EB009352. Longitudinal developing infant brain T1w/T2w MRI were provided by the Infant Brain Imaging Study (IBIS) Network, which is an NIH-funded Autism Centers of Excellence (ACE) project and consists of a consortium of 10 universities in the U.S. and Canada. 

\clearpage

\bibliographystyle{plain}
\bibliography{egbib}

\section*{Checklist}

\begin{enumerate}

\item For all authors...
\begin{enumerate}
  \item Do the main claims made in the abstract and introduction accurately reflect the paper's contributions and scope?
    \answerYes{See abstract and Sec.\ref{sec:introduction}}
  \item Did you describe the limitations of your work?
    \answerYes{See Sec.~\ref{sec:discussion}.}
  \item Did you discuss any potential negative societal impacts of your work?
    \answerYes{See Sec.~\ref{sec:discussion}.}
  \item Have you read the ethics review guidelines and ensured that your paper conforms to them?
    \answerYes{}
\end{enumerate}

\item If you are including theoretical results...
\begin{enumerate}
  \item Did you state the full set of assumptions of all theoretical results?
    \answerNA{No new theoretical results claimed.}
        \item Did you include complete proofs of all theoretical results?
    \answerNA{}
\end{enumerate}

\item If you ran experiments...
\begin{enumerate}
  \item Did you include the code, data, and instructions needed to reproduce the main experimental results (either in the supplemental material or as a URL)?
    \answerYes{Code: code with instructions is available at \url{https://github.com/mengweiren/longitudinal-representation-learning}. Data: \revise{public data links and procedures are described in the supplementary material}.}
  \item Did you specify all the training details (e.g., data splits, hyperparameters, how they were chosen)?
    \answerYes{See `Implementation details' in Sec.~\ref{sec:experiments} and App.~\ref{app:addn_impl_dets}.}
    \item Did you report error bars (e.g., with respect to the random seed after running experiments multiple times)?
    \answerYes{See Fig.~\ref{fig:result_boxplot}, Tab.~\ref{tab:result_moremetrics}, Tab.~\ref{tab:ablation}}.
        \item Did you include the total amount of compute and the type of resources used (e.g., type of GPUs, internal cluster, or cloud provider)?
    \answerYes{See `Implementation details' in Sec.~\ref{sec:experiments}}.
\end{enumerate}

\item If you are using existing assets (e.g., code, data, models) or curating/releasing new assets...
\begin{enumerate}
  \item If your work uses existing assets, did you cite the creators?
    \answerYes{See `Baselines and Evaluation Strategies' in Sec.~\ref{sec:experiments}}
  \item Did you mention the license of the assets?
    \answerNA{}
  \item Did you include any new assets either in the supplemental material or as a URL?
    \answerYes{Our code is available freely at \url{https://github.com/mengweiren/longitudinal-representation-learning}.}
  \item Did you discuss whether and how consent was obtained from people whose data you're using/curating?
    \answerNA{See~\cite{lamontagne2019oasis} and~\cite{shen2022subcortical} for how imaging consent was acquired for OASIS3 and IBIS, respectively.}
  \item Did you discuss whether the data you are using/curating contains personally identifiable information or offensive content?
    \answerYes{See `Data and preprocessing' in Sec.~\ref{sec:experiments}. Skull-stripping deidentifies human faces as only the brain is available.}
\end{enumerate}

\item If you used crowdsourcing or conducted research with human subjects...
\begin{enumerate}
  \item Did you include the full text of instructions given to participants and screenshots, if applicable?
    \answerNA{}
  \item Did you describe any potential participant risks, with links to Institutional Review Board (IRB) approvals, if applicable?
    \answerYes{See App. ~\ref{app:IRB}.}
  \item Did you include the estimated hourly wage paid to participants and the total amount spent on participant compensation?
    \answerNA{See App. ~\ref{app:more_data_details}.}
\end{enumerate}

\end{enumerate}

\newpage
\appendix

\clearpage
\section{Additional Results and Experiments}  \label{app:more_results}
\begin{figure}[!hb]
    \centering
    \includegraphics[width=0.95\textwidth]{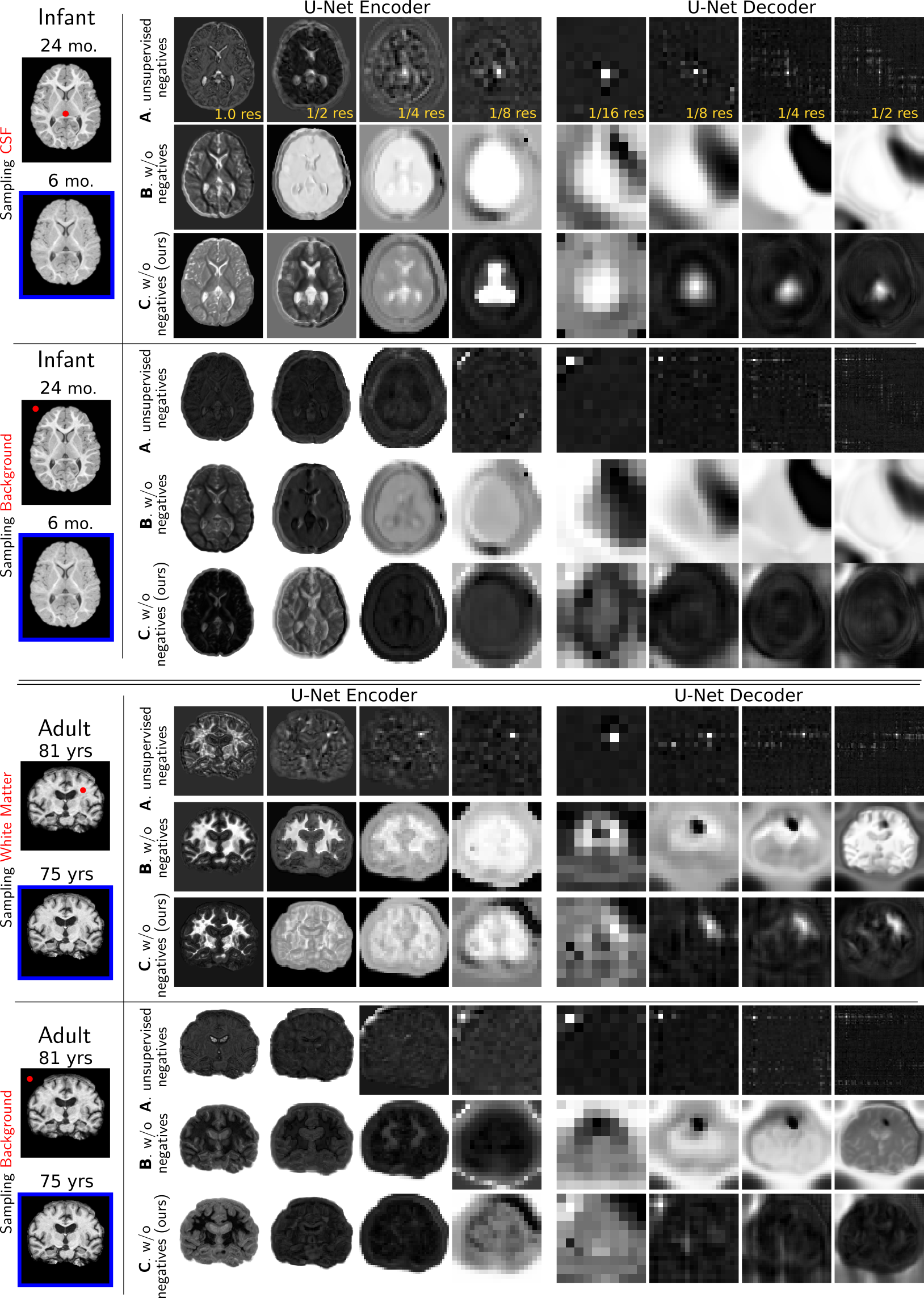}
    \caption{\textbf{Self-supervised spatiotemporal multi-scale similarity learning} visualized by sampling a spatial \textcolor{red}{query} point at given time point and computing its featurewise similarity to all locations in the \textcolor{blue}{key} image acquired at a different age. The similarity-maps obtained from both datasets reveal semantically-relevant anatomical representation within the U-Net and projector/predictor networks. This figure follows the same nomenclature (A--C) as Fig.~\ref{fig:simmap} with analysis in Sec.~\ref{sec:experiments} (main text).}
    \label{fig:simmap_more}
\end{figure}

\textbf{Self-supervised similarity maps} from anatomically-relevant key points are visualized in Figure \ref{fig:simmap_more}. Using only self-supervision, our model learns semantically and positionally-aware representations. 

\begin{figure}[!h]
    \centering
    \includegraphics[width=\textwidth]{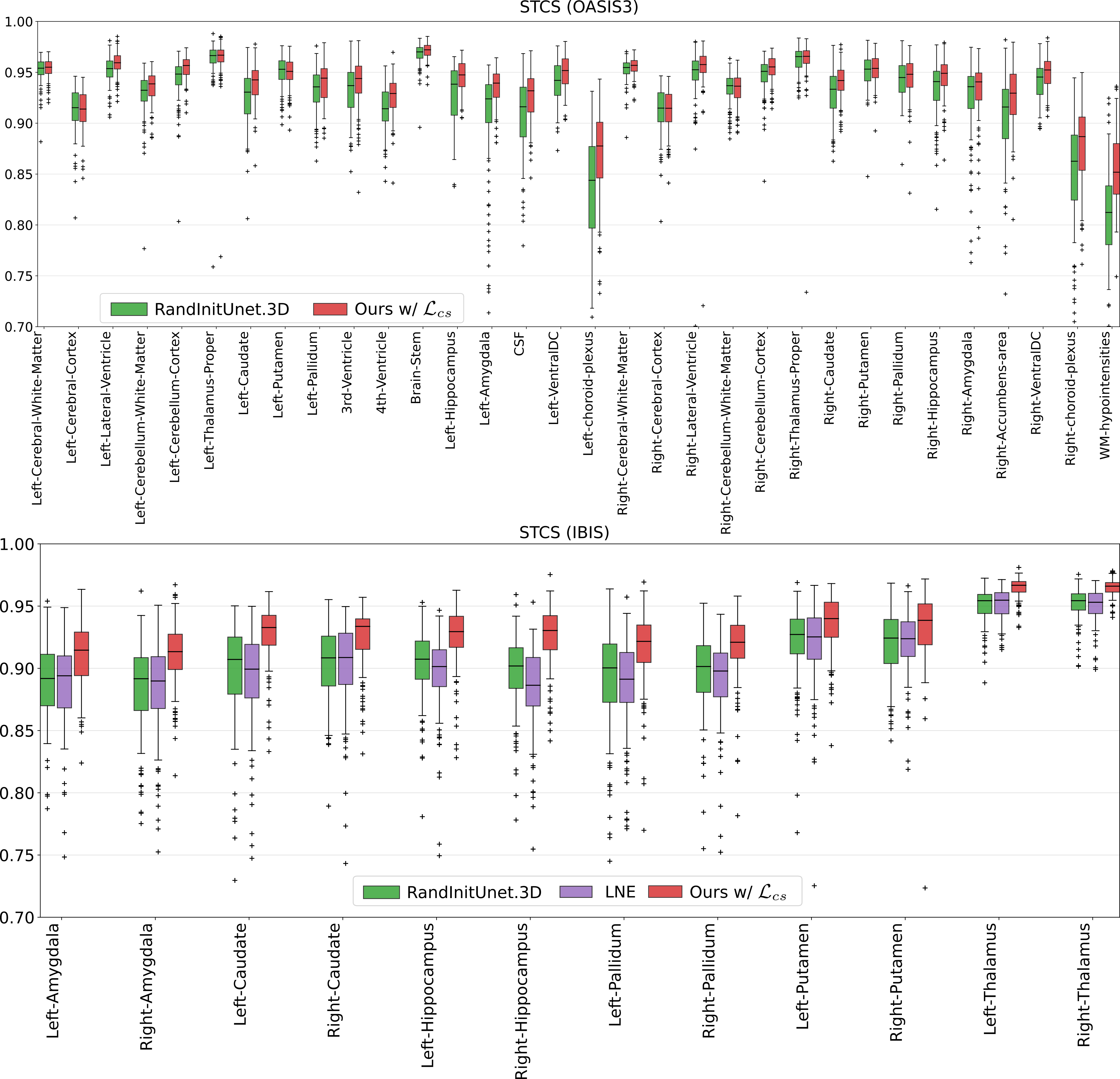}
    \caption{\textbf{Per-label longitudinal consistency comparison between the best performing models} as measured by STCS (spatiotemporal consistency of segmentations~\cite{li2021longitudinal}, higher is better) in the one-shot segmentation setting. For space considerations, only the best-performing two and three models are visualized on test set for the OASIS3 and IBIS-\textit{subcort} tasks, respectively.}
    \label{fig:STCS_per_label}
\end{figure}

\textbf{Label-wise longitudinal consistency.}
To observe structure-specific performance, we report label-wise longitudinal consistency scores (as measured by STCS~\cite{li2021longitudinal}) for OASIS3 and IBIS-\textit{subcort} on the test set for the best performing models (as measured by Figure \ref{fig:result_boxplot} of the main text) in Figure \ref{fig:STCS_per_label}. IBIS-\textit{wmgm} was not included as it only has two anatomical structures.

\textbf{Few-shot and fully-supervised segmentation.}
While the main body of the paper presents results on one-shot segmentation (corresponding to the common real-world setting of single atlas-based segmentation), the proposed framework has benefits beyond one-shot segmentation. We present results obtained by using 10\% and 100\% of all the labeled training sets in Tables \ref{tab:finetune_0.1} and \ref{tab:finetune_1.0}, respectively. As having more supervised data may necessitate changing our hyperparameters, we explore reducing the weight of the consistency-regularization from $1.0$ to $0.1$ for both experiments. In the 10\% setting, we find that \texttt{Ours w/ }$\mathcal{L}_{cs}$ with a weight of $0.1$ obtains optimal performance and consistency. In the 100\% setting, we find that including $\mathcal{L}_{cs}$ actually degrades performance and that optimal results are achieved with just the proposed pretraining (\texttt{Ours w/o }$\mathcal{L}_{cs}$). We hypothesize that these trends arise from imperfect deformable registration of the intra-subject images as they were warped considering only their intensity and not semantic structure (see App Section \ref{app:registration}). Therefore, having the imperfect self-supervision of $\mathcal{L}_{cs}$ in the low-annotation regime increases performance. However, hundreds of annotated samples remove the need for $\mathcal{L}_{cs}$, while maintaining the segmentation performance benefits of our pretraining framework over existing work. 

\begin{table}[ht]
    \footnotesize
    \centering
    \caption{\textbf{\textit{Few}-shot segmentation} using $\mathbf{10\%}$ of the annotated training set for finetuning. This table quantifies test set performance (Dice, IoU, HD95) and longitudinal consistency (ASPC~\cite{reuter2012within}, STCS\cite{li2021longitudinal}). Mdn: Median.}
    \begin{tabular}{cccccccc}
    \multicolumn{8}{c}{IBIS-subcort} \\ \toprule
    Model                        & Mean(std) Dice  & Mdn Dice & Mean(std) IoU    & Mdn HD95  & Mdn IoU & ASPC & STCS\\ \midrule
    RandInitUnet.2D              & 0.895(0.03) &  0.891 &  0.812(0.06)  &  2.390 & 0.804 & 4.618 & 0.909 \\ 
    RandInitUnet.3D              & 0.909(0.03) &  0.905 &  0.834(0.05)  &  1.819 & 0.827 & 5.531 & 0.911 \\ 
    Context Restore~\cite{chen2019self} & 0.901(0.03) &  0.898 &  0.821(0.05)  &  2.240 & 0.815 & 4.895 & 0.918\\ 
    LNE~\cite{ouyang2021self} & 0.914(0.03) &  0.909 &  0.842(0.05)  &  1.845 & 0.834 & 5.807 & 0.911\\ 
    GLCL~\cite{chaitanya2020contrastive} & 0.905(0.03) &  0.900 &  0.827(0.05)  &  2.486 & 0.819 & 4.369 & 0.913\\ 
    PCL~\cite{zeng2021positional} & 0.904(0.03) &  0.900 &  0.826(0.05)  &  2.881 & 0.819 & 4.479  & 0.911\\ 
    PatchNCE~\cite{park2020contrastive} & 0.919(0.03) &  0.915 &  0.851(0.05)  &  1.004 & 0.844 & 5.871 & 0.908\\ \hline
    Ours w/o $\mathcal{L}_{cs}$  & 0.920(0.03) &  0.916 &  0.853(0.05)  &  \textbf{1.001} & 0.846 & 5.521 & 0.908\\ 
    Ours w/ $0.1\mathcal{L}_{cs}$& \textbf{0.923(0.03)} &  \textbf{0.920} &  \textbf{0.858(0.04)}  &  1.002 & \textbf{0.853} & 5.077 & 0.910\\
    Ours w/ $1\mathcal{L}_{cs}$  & 0.917(0.03) &  0.913 &  0.848(0.05)  &  1.002 & 0.840 & \textbf{4.655} & \textbf{0.925} \\
    \bottomrule
    \multicolumn{8}{c}{}\\
    \multicolumn{8}{c}{OASIS3} \\    \toprule
    Model                 & Mean(std) Dice  & Mdn Dice & Mean(std) IoU    & Mdn HD95  & Mdn IoU & ASPC & STCS\\ \midrule
    RandInitUnet.2D              & 0.852(0.09) &  0.878 &  0.752(0.13)  &  1.569 & 0.785  & 4.297 & 0.908 \\  
    RandInitUnet.3D              & 0.867(0.09) &  0.888 &  0.774(0.12)  &  1.409 & 0.800  & 2.575 & 0.926 \\  
    Context Restore~\cite{chen2019self} & 0.847(0.10) &  0.873 &  0.745(0.13)  &  1.660 & 0.777  & 4.528 & 0.916\\  
    LNE~\cite{ouyang2021self} & 0.872(0.09) &  0.894 &  0.782(0.12)  &  1.426 &\textbf{ 0.811}  & 2.747 & 0.923\\  
    GLCL~\cite{chaitanya2020contrastive} & 0.861(0.09) &  0.884 &  0.765(0.12)  &  1.387 & 0.794  & 4.005 & 0.911\\ 
    PCL~\cite{zeng2021positional} & 0.862(0.09) &  0.885 &  0.766(0.12)  &  1.328 & 0.796  & 3.896 & 0.910\\ 
    PatchNCE~\cite{park2020contrastive} & 0.868(0.09) &  0.888 &  0.776(0.12)  &  1.347 & 0.800  & 2.666 & 0.927\\ \hline
    Ours w/o $\mathcal{L}_{cs}$  & 0.872(0.08) &  0.894 &  0.782(0.12)  &  1.372 & 0.810  & 2.647 & 0.923\\
    Ours w/ $0.1\mathcal{L}_{cs}$ &\textbf{ 0.873(0.08)} &  \textbf{0.894} &  \textbf{0.783(0.12)}  &  \textbf{1.335} & 0.809  & 2.602 & 0.926\\
    Ours w/ $1\mathcal{L}_{cs}$   & 0.868(0.09) &  0.890 &  0.776(0.12)  &  1.341 & 0.804  & \textbf{2.446} & \textbf{0.932}\\
    \bottomrule
    \end{tabular}
    \label{tab:finetune_0.1}
\end{table}

\begin{table}[h]
    \footnotesize
    \centering
    \caption{\textbf{\textit{Fully}-supervised segmentation} using $\mathbf{100\%}$ of the annotated training set for finetuning. This table quantifies test set performance (Dice, IoU, HD95) and longitudinal consistency (ASPC~\cite{reuter2012within}, STCS\cite{li2021longitudinal}). Mdn: Median.}
    \begin{tabular}{cccccccc}
    \multicolumn{8}{c}{IBIS-subcort} \\ \toprule
    Model                       & Mean(std) Dice  & Mdn Dice & Mean(std) IoU    & Mdn HD95  & Mdn IoU & ASPC & STCS\\ \midrule
    RandInitUnet.2D              & 0.902(0.03) &  0.898 &  0.824(0.05)  &  2.634 & 0.815  & 4.308 & 0.917 \\  
    RandInitUnet.3D              & 0.922(0.03) &  0.917 &  0.856(0.04)  &  1.730 & 0.847  & 5.517 & 0.916 \\  
    Context Restore~\cite{chen2019self} & 0.909(0.03) &  0.906 &  0.835(0.05)  &  1.011 & 0.828  & 4.125 & 0.922\\  
    LNE~\cite{ouyang2021self} & 0.928(0.02) &  0.924 &  0.867(0.04)  &  1.000 & 0.859  & 5.253 & 0.911\\  
    GLCL~\cite{chaitanya2020contrastive} & 0.914(0.03) &  0.909 &  0.843(0.05)  &  1.011 & 0.834  & 4.191 & 0.917 \\ 
    PCL~\cite{zeng2021positional} & 0.913(0.03) &  0.909 &  0.842(0.05)  &  1.006 & 0.833  & \textbf{4.122} & 0.918\\  
    PatchNCE~\cite{park2020contrastive} & 0.928(0.02) &  0.924 &  0.867(0.04)  &  1.001 & 0.860  & 5.470 & 0.910\\ \hline
    Ours w/o $\mathcal{L}_{cs}$  & \textbf{0.933(0.02)} &  \textbf{0.930} & \textbf{ 0.876(0.04)}  &  1.000 & \textbf{0.870}  & 6.666 & 0.901 \\  
    Ours w/ $0.1\mathcal{L}_{cs}$ & 0.930(0.02) &  0.926 &  0.870(0.04)  &  1.000 & 0.863  & 5.298 & 0.910\\
    Ours w/ $1\mathcal{L}_{cs}$   & 0.922(0.02) &  0.917 &  0.856(0.04)  &  1.000 & 0.846  & 4.605 & \textbf{0.926}\\  
    \bottomrule
    \multicolumn{8}{c}{}\\
    \multicolumn{8}{c}{OASIS3} \\    \toprule
    Model                 & Mean(std) Dice  & Mdn Dice & Mean(std) IoU    & Mdn HD95  & Mdn IoU & ASPC & STCS\\ \midrule
    RandInitUnet.2D               & 0.844(0.10) &  0.874 &  0.741(0.13)  &  1.632 & 0.777  & 4.438 & 0.914\\ 
    RandInitUnet.3D               & 0.878(0.08) &  0.898 &  0.791(0.11)  &  1.371 & 0.817  & 5.570 & 0.920 \\ 
    Context Restore~\cite{chen2019self} & 0.864(0.09) &  0.886 &  0.769(0.12)  &  1.415 & 0.797  & 4.202 & 0.927\\ 
    LNE~\cite{ouyang2021self}& 0.882(0.08) &  0.904 &  0.796(0.11)  &  1.280 & 0.825  & 4.140 & 0.926\\  
    GLCL~\cite{chaitanya2020contrastive} & 0.864(0.09) &  0.887 &  0.769(0.12)  &  1.424 & 0.799  & 3.590 & 0.926 \\ 
    PCL~\cite{zeng2021positional} & 0.865(0.09) &  0.890 &  0.771(0.12)  &  1.315 & 0.802  & 3.328 & 0.926\\ 
    PatchNCE~\cite{park2020contrastive}& 0.869(0.08) &  0.889 &  0.777(0.11)  &  1.328 & 0.801  & 2.599 & 0.937\\ \hline
    Ours w/o $\mathcal{L}_{cs}$   & \textbf{0.885(0.08)} &  \textbf{0.907} &  \textbf{0.801(0.11)}  &  1.246 & \textbf{0.831}  & 2.919 & 0.933\\ 
    Ours w/ $0.1\mathcal{L}_{cs}$ & 0.882(0.08) &  0.904 &  0.796(0.11)  &  \textbf{1.233} & 0.825  & 2.647 & 0.934\\
    Ours w/ $1\mathcal{L}_{cs}$   & 0.877(0.08) &  0.898 &  0.789(0.11)  &  1.245 & 0.817  & \textbf{2.368} & \textbf{0.939}\\
    \bottomrule
    \end{tabular}
    \label{tab:finetune_1.0}
\end{table}

\clearpage

\textbf{U-Net configuration.} 
For consistency, we use the same base U-Net architecture for all baselines. Its configuration is modeled based on the one-shot segmentation mean Dice validation results on OASIS3 presented in Table \ref{tab:unet_specs} and evaluated over several training crop sizes, normalization layers, and channel width multipliers. Importantly, all configurations were trained without any pretraining to obtain a baseline. We observe that a random crop window of $128^3$ is optimal (rows A, B, C) and find that Instance Normalization (row D) instead of Batch Normalization degrades performance. Finally, we observe almost no performance degradation when reducing the model size via the channel width multiplier from 24 to 16 (row C vs E) and therefore use 16 for computational efficiency.

\begin{table}[h!]
\centering
\caption{\textbf{Base network tuning.} Randomly initialized UNet performance over various crop sizes, normalization types, and the \# channels of the first convolution layer. E is the final configuration for the base network used in all baselines as it best trades-off memory usage and performance.}
\begin{tabular}{ccccc}
\toprule
Exp & Crop size & Normalization & Channel width & mean(std) dice     \\ \midrule
A   & $64 \times 64 \times 64$          & batch norm         & 24 & 0.768 ($\pm$ 0.14) \\
B   & $128 \times 128 \times 128$       & batch norm         & 24 & \textbf{0.795 ($\pm$ 0.14)} \\
C   & $160\times160\times192$           & batch norm         & 24 & 0.786 ($\pm$ 0.14) \\
D   & $128 \times 128 \times 128$       & instance norm      & 24 & 0.777 ($\pm$ 0.14) \\
E   & $128 \times 128 \times 128$       & batch norm         & 16 & 0.792 ($\pm$ 0.15) \\
F   & $128 \times 128 \times 128$       & batch norm         & 8 & 0.786 ($\pm$ 0.14)\\
\bottomrule
\end{tabular}
\label{tab:unet_specs}
\end{table}

\begin{figure}[h!]
    \centering
    \includegraphics[width=0.45\textwidth]{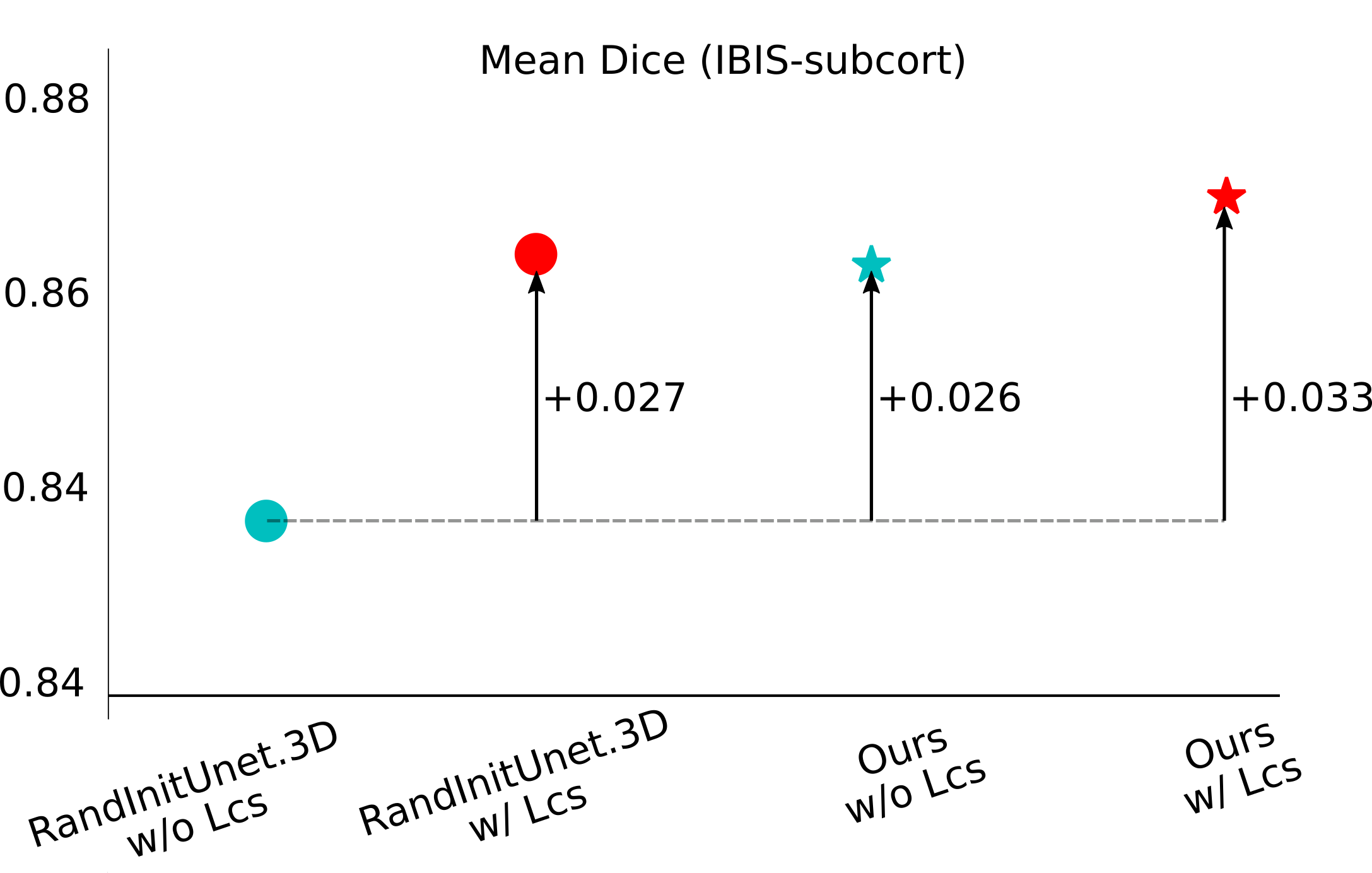}
    \caption{Stand-alone combination of the proposed $\mathcal{L}_{cs}$ consistency-regularization and one-shot segmentation training of a randomly-initialized U-Net. In the one-shot setting, $\mathcal{L}_{cs}$ improves mean dice by 0.027, which is comparable to the proposed model without $\mathcal{L}_{cs}$. Optimal results are obtained when combined with our regularized pretraining framework.}
    \label{fig:Lcs}
\end{figure}

\textbf{Training a randomly-initialized U-Net with $\mathcal{L}_{cs}$.} To investigate the standalone benefit of $\mathcal{L}_{cs}$ without any regularized pretraining, we apply $\mathcal{L}_{cs}$ to a randomly initialized 3D U-Net in the one-shot segmentation setting for IBIS-\textit{subcort} and report its results in Fig.~\ref{fig:Lcs}. 

\begin{figure}[!hb]
    \centering
    \includegraphics[width=0.4\textwidth]{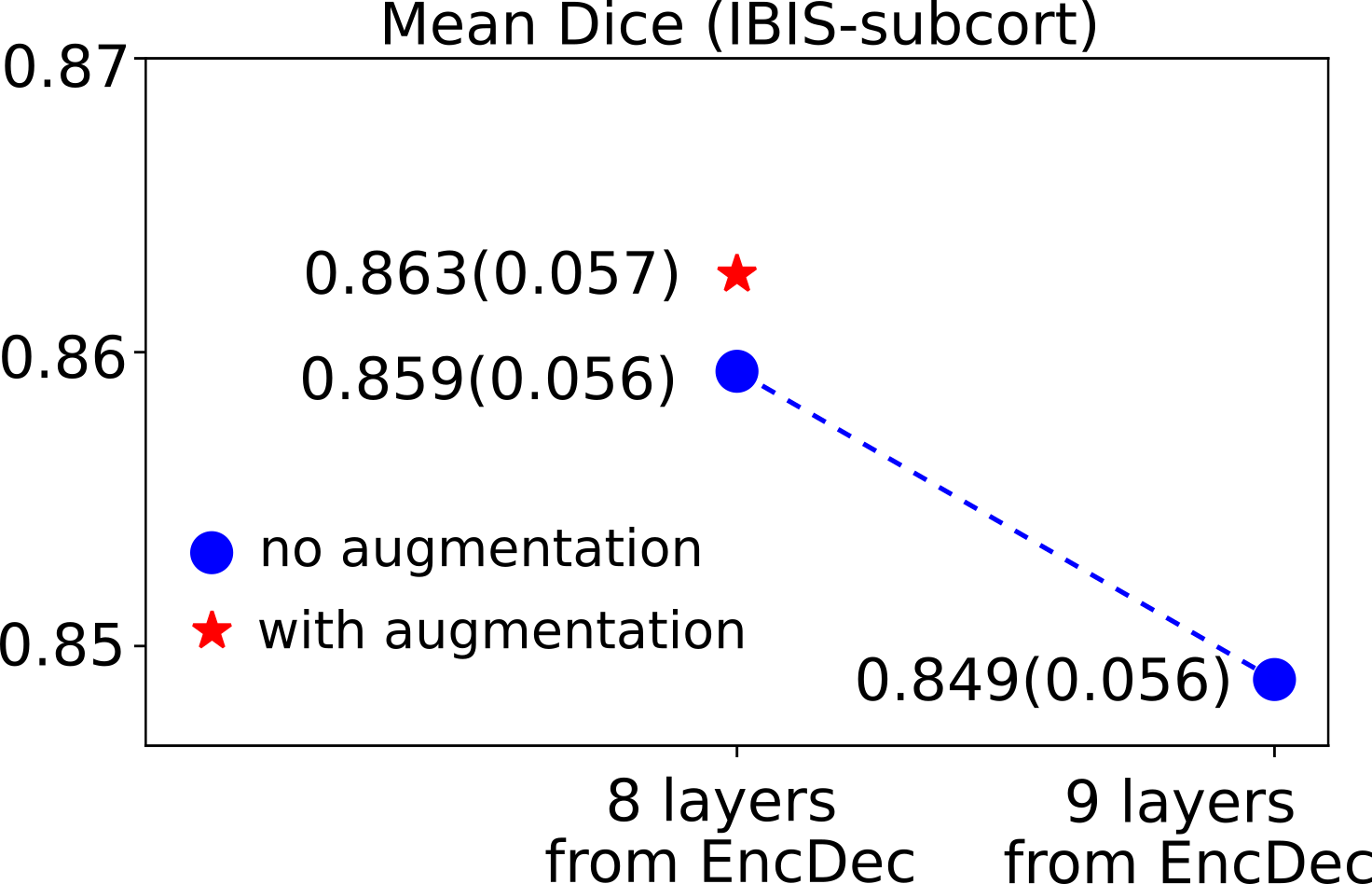}
    \caption{A comparison between using different number of layers during pretraining. We used `8 layers from EncDec' as our final layer selection as a trade-off between memory and performance.}
    \label{fig:nlayers}
\end{figure}

\textbf{Using last decoder layer in loss.} We investigate the impact of including the full-resolution feature (layer 21 in Table~\ref{tab:unet_arch}) in the \texttt{EncDec} loss configuration described in Section \ref{app:addn_training_dets}. As shown in Figure \ref{fig:nlayers}, this notably degrades performance, indicating sensitivity to the exact layers used for $\mathcal{L}_{sim}$ calculation.

\textbf{Additional modeling decisions.}
In addition to the quantitative ablations over our modeling choices included in the main paper and appendix, we make a few modeling decisions detailed in Table \ref{tab:param} based on qualitative assessments of the representation quality using feature similarity maps. 
\begin{table}[h]
    \centering
        \caption{Additional modeling decisions. The final selections are bolded.}
    \footnotesize
    \begin{tabular}{cc}
       \toprule
       \textbf{Parameter}  & \textbf{Search space} \\ \midrule
       \# of layers in projector  & 2,\textbf{3} \\
       non-linearity & ELU, \textbf{ReLU}, LeakyReLU \\
       within subject registration & affine only, \textbf{affine and deformable} \\
       feature selection for similarity loss & \textbf{conv output}, activation output \\ \bottomrule
    \end{tabular}
    \label{tab:param}
\end{table}

\looseness=-1
\textbf{Additional one-shot quantification.} As Table~\ref{tab:result_moremetrics} and Figure~\ref{fig:result_boxplot} (main text) report median scores, Table \ref{tab:one_shot_mean_std} reports means and standard deviations for further interpretation. Similar to the medians, our method improves both mean performance and longitudinal consistency on IBIS-\{\textit{wmgm, subcort}\}. On OASIS3, our Dice and IoU performance improves on all baselines except for LNE (with whom it is strongly competitive) and exceeds all baselines in terms of longitudinal consistency (ASPC, STCS).

\begin{table}[h]
    \footnotesize
    \centering
    \caption{\textbf{One-shot segmentation performance} to complement results presented in Tab.~\ref{tab:result_moremetrics} and Fig.~\ref{fig:result_boxplot} of the main text, using means instead of medians. *In OASIS3, we exclude the \texttt{Right Accumbens Area} and \texttt{4th Ventricle} labels when calculating mean HD95 due to numerical issues for some of the baselines (\texttt{RandInitUnet.2D}, \texttt{Context Restore}, and \texttt{PCL}).}
    \begin{tabular}{cccccc}
    \multicolumn{6}{c}{IBIS-subcort} \\ \toprule
    Model                 & Dice  &  IoU   & HD95     & ASPC & STCS \\
    \midrule
    RandInitUnet.2D              & 0.828(0.07) &  0.712(0.10) &  4.784(6.32) & 7.08(7.85) & 0.893(0.03)\\
    RandInitUnet.3D              & 0.836(0.06) &  0.724(0.09) &  3.603(5.27) & 5.29(5.12) & 0.908(0.02)\\
    Context Restore~\cite{chen2019self} & 0.838(0.07) &  0.727(0.09) &  5.196(9.27) & 8.22(9.16) & 0.893(0.02)\\
    LNE~\cite{ouyang2021self} & 0.847(0.06) &  0.739(0.09) &  3.478(5.33) & 5.83(5.41) & 0.905(0.02)\\
    GLCL~\cite{chaitanya2020contrastive} & 0.835(0.07) &  0.722(0.10) &  3.640(4.16) & 5.47(6.23) & 0.905(0.03)\\
    PCL~\cite{zeng2021positional} & 0.833(0.07) &  0.720(0.09) &  3.528(3.56) & 5.78(6.10) & 0.902(0.03)\\
    PatchNCE~\cite{park2020contrastive} & 0.856(0.06) &  0.753(0.09) &  1.482(1.75) & 5.93(5.79) & 0.909(0.02)\\
    Ours w/o $\mathcal{L}_{cs}$  & 0.863(0.06) &  0.763(0.08) &  1.510(2.62) & 5.68(6.00) & 0.909(0.02)\\
    Ours w/ $\mathcal{L}_{cs}$   & \textbf{0.870(0.05)} & \textbf{ 0.773(0.08) }&  \textbf{1.408(1.24)} & \textbf{4.26(4.32)} & \textbf{0.929(0.02)}\\
    \bottomrule
    \multicolumn{6}{c}{}\\
    \multicolumn{6}{c}{IBIS-wmgm} \\ \toprule
    Model                 & Dice  &  IoU   & HD95     & ASPC & STCS \\ \midrule
    RandInitUnet.2D              & 0.672(0.07) &  0.510(0.08) &  3.274(1.00) & 11.51(11.79) & 0.832(0.03)\\
    RandInitUnet.3D              & 0.713(0.08) &  0.560(0.09) &  3.788(0.70) & 5.52(5.48)   & 0.863(0.05)\\
    Context Restore~\cite{chen2019self} & 0.590(0.19) &  0.444(0.19) &  8.273(3.56) & 29.24(26.76) & 0.813(0.10)\\
    LNE~\cite{ouyang2021self} & 0.716(0.07) &  0.563(0.09) &  \textbf{3.201(0.98)} & 5.35(6.34)   & 0.858(0.04)\\
    GLCL~\cite{chaitanya2020contrastive} & 0.707(0.05) &  0.550(0.06) &  4.112(1.14) & 8.47(8.85)   & 0.858(0.02)\\
    PCL~\cite{zeng2021positional} & 0.715(0.08) &  0.562(0.09) &  4.974(1.75) & 10.65(13.29) & 0.868(0.05)\\
    PatchNCE~\cite{park2020contrastive} & 0.753(0.05) &  0.607(0.06) &  4.344(0.66) & 3.78(4.18)   & 0.889(0.03)\\
    Ours w/o $\mathcal{L}_{cs}$  & 0.758(0.06) &  0.614(0.08) &  3.291(0.66) & \textbf{2.46(2.74)}   & 0.882(0.03)\\
    Ours w/ $\mathcal{L}_{cs}$   & \textbf{0.806(0.03)} &  \textbf{0.676(0.04)} &  3.237(0.47) & 4.15(3.32)   & \textbf{0.910(0.02)}\\ \bottomrule
    \multicolumn{6}{c}{}\\
    \multicolumn{6}{c}{OASIS3} \\ \toprule
    Model                & Dice  &  IoU   & HD95*     & ASPC & STCS \\
     \midrule
    RandInitUnet.2D              & 0.781(0.15) &  0.661(0.17) &  3.425(5.00)     & 10.27(28.84) & 0.873(0.05)\\
    RandInitUnet.3D              & 0.804(0.13) &  0.689(0.16) &  3.424(5.85)     & 3.53(6.35)   & 0.923(0.05)\\
    Context Restore~\cite{chen2019self} & 0.801(0.14) &  0.687(0.16) &  4.841(10.00)    & 6.88(15.21)  & 0.899(0.04)\\
    LNE~\cite{ouyang2021self} & $\textbf{0.813(0.13)}$ &  0.700(0.15) &  \textbf{3.171(5.50)}     & 8.81(31.94)  & 0.898(0.04)\\
    GLCL~\cite{chaitanya2020contrastive} & 0.792(0.14) &  0.674(0.16) &  3.330(5.25)     & 5.30(11.12)  & 0.903(0.05)\\
    PCL~\cite{zeng2021positional} & 0.796(0.14) &  0.680(0.17) &  3.339(5.41)     & 5.69(13.40)  & 0.899(0.05)\\
    PatchNCE~\cite{park2020contrastive} & 0.809(0.13) &  0.697(0.16) &  3.298(5.93)     & 3.49(6.70)   & 0.920(0.04)\\
    Ours w/o $\mathcal{L}_{cs}$                 & \textbf{0.813(0.13)} &  \textbf{0.702(0.15)} &  3.289(5.68)     & 3.40(6.11)   & 0.923(0.04)\\
    Ours w/ $\mathcal{L}_{cs}$                  & 0.810(0.13) &  0.697(0.16) &  3.733(6.41)     & \textbf{2.76(4.96) }  & \textbf{0.934(0.03)}\\
    \bottomrule
    \end{tabular}
    \label{tab:one_shot_mean_std}
\end{table}

\textbf{Additional IBIS-\textit{wmgm} segmentation results.} For one-shot IBIS-\textit{wmgm} segmentation, we train on a single 36 month old T1w/T2w MR image and quantitatively evaluate on 6 month old MR images (for additional motivation see Appendix Section \ref{app:wmgm_motivation}). In Figure \ref{fig:addn_wmgm_results}, we visualize predictions using additional baselines and our framework on a held-out test subject. As the infant brain matures over time (Appendix Section \ref{app:wmgm_motivation}), the rows are arranged in order of the most to the least amount of biological domain-shift w.r.t. the 36 month-old training image. Further, as the 12, 24, and 36 month old held-out images do not have ground truth segmentations available, we are restricted to qualitative evaluations of segmentation quality.

\begin{figure}[h]
    \centering
    \includegraphics[width=\textwidth]{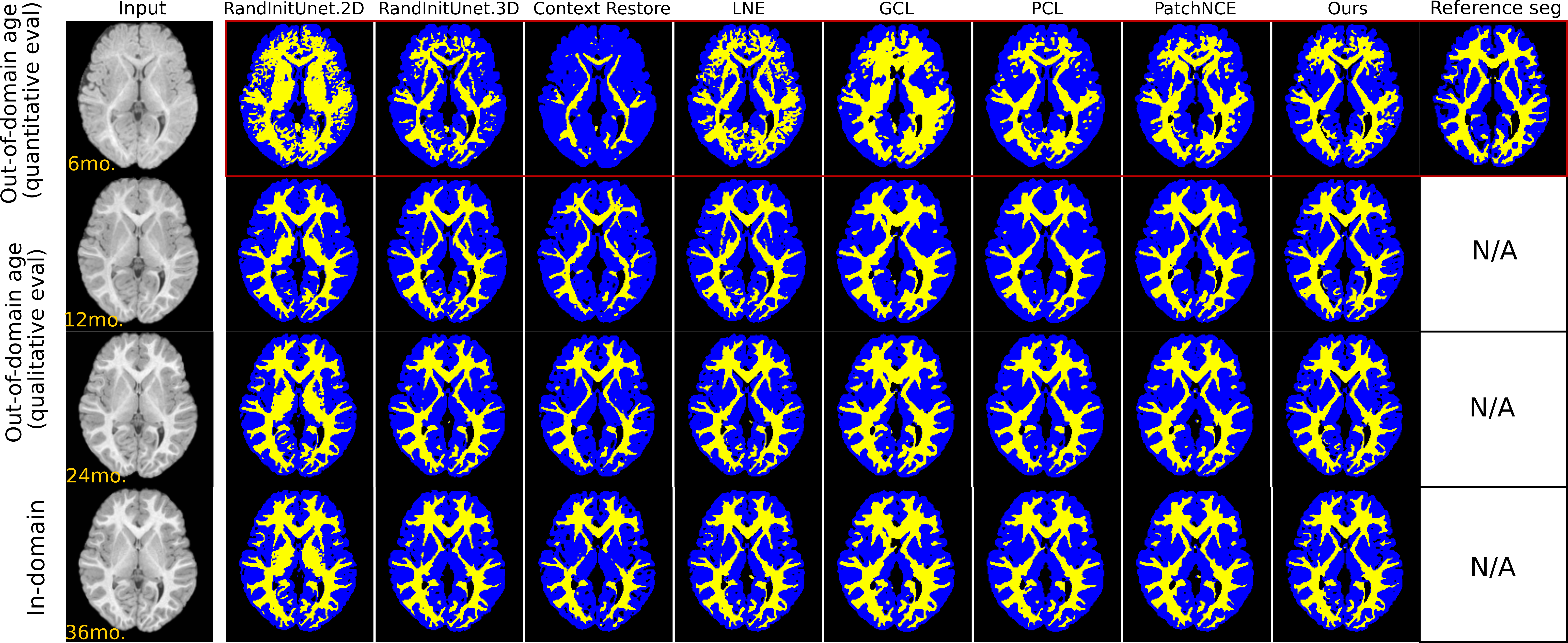}
    \caption{Infant brain segmentation on a held-out subject under the strong domain shift induced by training on a single 36 month-old image and testing on the remaining timepoints. Under the most extreme domain shift of row 1, we see that all baselines yield highly-inaccurate predictions, while our method yields higher segmentation quality.}
    \label{fig:addn_wmgm_results}
\end{figure}

\clearpage
\section{Additional Implementation Details}\label{app:addn_impl_dets}

\subsection{Additional Training Details} \label{app:addn_training_dets}
\begin{figure}[ht]
    \centering
    \includegraphics[width=\textwidth]{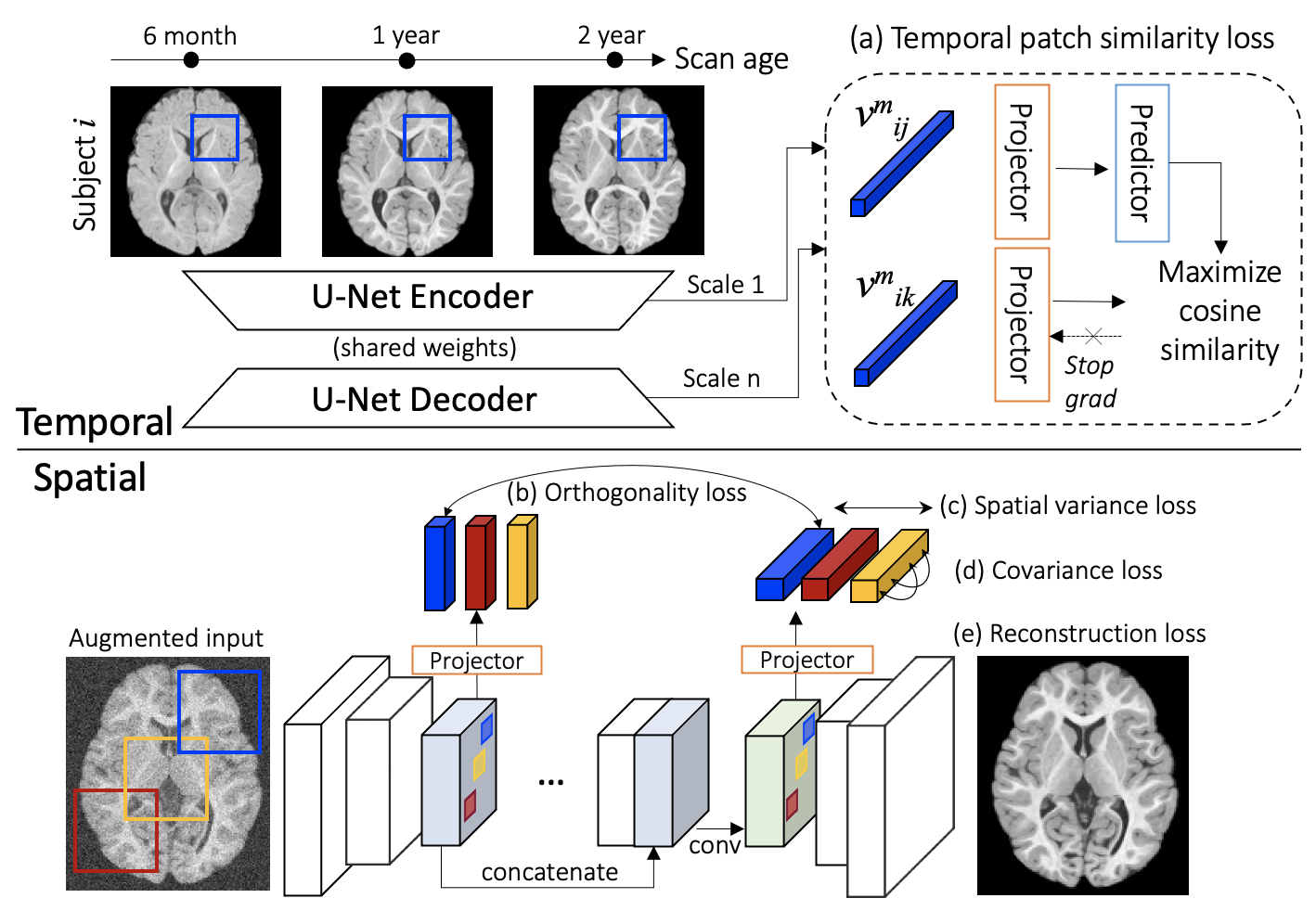}
    \caption{\textbf{Pre-training Overview.} A visual illustration of the proposed losses (a, b, c, d, e) for spatiotemporal representation learning with a U-Net. An overview of (a)--(d) is in main text Fig.1.}
    \label{fig:detailed_overview}
\end{figure}

Figure \ref{fig:detailed_overview} illustrates an in-depth overview of the proposed pretraining and representation learning framework. In the proposed loss configuration (\texttt{EncDec}), Pre-activation features from layers $\{1,3,5,7,9,12,15,18\}$ in Table \ref{tab:unet_specs} are sampled spatially to extract channel-wise vectors which are fed into the projector and predictor MLPs for $\mathcal{L}_{sim}$ calculation. Patch features from layer $\{8,12\}$ in the bottleneck (processed by the projector) are used for the orthogonality regularizer $\mathcal{L}_{O}$ and layers $\{12,15,18\}$ in the decoder are used for the $\mathcal{L}_{S}$ and $\mathcal{L}_{C}$ regularizers. In the ablations in Table \ref{tab:ablation} (main text), the \texttt{Enc} loss setting refers to using features from layers $\{1,2,3,4,5,6,7,8,9,10,11\}$ for $\mathcal{L}_{sim}$.
Each iteration of pretraining loads three corresponding crops 
from two intra-subject images each for $\mathcal{L}_{sim}$ calculation. Higher batch sizes could have been used for the proposed loss functions but were not for consistency with one of our baselines (\texttt{PatchNCE}~\cite{park2020contrastive}) which is highly memory-intensive due to the need for a large number of negative samples alongside 3D computation.

During finetuning, the segmentation loss ($\mathcal{L}_{sup}$) is applied to the original unwarped images and the consistency loss ($\mathcal{L}_{cs}$) is applied to the nonlinearly aligned images. When calculating $\mathcal{L}_{sup}$ and $\mathcal{L}_{cs}$, we perform individual forward/backward passes for each loss due to memory limitations as gradient accumulation over two forward passes was found to decrease performance on a validation set.

\subsection{Architectures}
The MLP architectures for the \textbf{projector} and \textbf{predictor} networks used for pretraining and representation learning are given in Table \ref{tab:projector_predictor}. The \textbf{U-Net} architecture used for both pretraining and finetuning is given in Table \ref{tab:unet_arch}. During pretraining, an additional convolutional layer is used following layer 23 to reconstruct/denoise the input. During finetuning, a channel-wise softmax layer is attached following layer 23 for segmentation. 

\begin{table}[h]
\centering
\caption{Projector and predictor MLP architectures. }
\begin{tabular}{lll}
\multicolumn{3}{c}{Projector (f)}                               \\\toprule
id & Layer              & Output size                           \\ \midrule
0  & FC(2048), BN, ReLU & (num patches $\times$ batch size) $\times$ 2048 \\
1  & FC(2048), BN, ReLU & (num patches $\times$ batch size) $\times$ 2048 \\
2  & FC(2048), BN       & (num patches $\times$ batch size) $\times$ 2048 \\ \bottomrule 
\multicolumn{3}{l}{}                               \\
\multicolumn{3}{c}{Predictor (p)}                               \\\toprule
id & Layer              & Output size                           \\ \midrule
0  & FC(256), BN, ReLU & (num patches $\times$ batch size) $\times$ 256 \\
1  & FC(2048), BN, ReLU & (num patches $\times$ batch size) $\times$ 2048 \\ \bottomrule
\end{tabular}
\label{tab:projector_predictor}
\end{table}

\begin{table}[!h]
\setlength{\tabcolsep}{11pt}
\centering
\caption{U-Net architecture. All convolutional layers use $3\times 3$ kernels. BN: Batch Normalization (using default PyTorch momentum). The batch size dimension is denoted $bs$ and $nc$ is the starting channel width multiplier of the model. We choose $nc = 16$ consistently throughout all models based on Table\ref{tab:unet_specs}. $n$ indicates the number of output channels and is set to the number of labels.}
\begin{tabular}{lll}
\toprule
id                      & Layer                             & Output size                 \\ \midrule
0                       & Conv3D(nc), BN, ReLU              & bs, w, h, d, nc \\ 
1                       & Conv3D(nc), BN, ReLU              & bs, w, h, d, nc \\ 
2                       & Conv3D(nc), BN, ReLU              & bs, w, h, d, nc \\ 
3                       & MaxPool(2), Conv3D(2nc), BN, ReLU  & bs, w/2, h/2, d/2, 2nc\\
4                       & Conv3D(2nc), BN, ReLU             & bs, w/2, h/2, d/2, 2nc \\ 
5                       & MaxPool(2), Conv3D(4nc), BN, ReLU & bs, w/4, h/4, d/4, 4nc \\
6                       & Conv3D(4nc), BN, ReLU             & bs, w/4, h/4, d/4, 4nc \\
7                       & MaxPool(2), Conv3D(8nc), BN, ReLU & bs, w/8, h/8, d/8, 8nc \\
8                       & Conv3D(8nc), BN, ReLU            & bs, w/8, h/8, d/8, 8nc \\
9                       & MaxPool(2), Conv3D(16nc), BN, ReLU & bs, w/16, h/16, d/16, 16nc \\
10                      & Conv3D(16nc), BN, ReLU            & bs, w/16, h/16, d/16, 16nc \\
11                      & Upsample(2), Concatenate with 8 & bs, w/8, h/8, d/8, 24nc \\
12                      & Conv3D(16nc), BN, ReLU            & bs, w/8, h/8, d/8, 8nc \\
13                      & Conv3D(16nc), BN, ReLU            & bs, w/8, h/8, d/8, 8nc \\
14                      & Upsample(2), Concatenate with 6 & bs, w/4, h/4, d/4, 12nc \\
15                      & Conv3D(4nc), BN, ReLU             & bs, w/4, h/4, d/4, 4nc \\
16                      & Conv3D(4nc), BN, ReLU             & bs, w/4, h/4, d/4, 4nc \\
17                      & Upsample(2), Concatenate with 4 & bs, w/2, h/2, d/2, 6nc \\
18                      & Conv3D(2nc), BN, ReLU             & bs, w/2, h/2, d/2, 2nc \\
19                      & Conv3D(2nc), BN, ReLU             & bs, w/2, h/2, d/2, 2nc \\
20                      & Upsample(2), Concatenate with 2 & bs, w, h, d, 3nc \\
21                      & Conv3D(nc), BN, ReLU             & bs, w, h, d, nc \\
22                      & Conv3D(nc), BN, ReLU             & bs, w, h, d, nc \\
23                      & Conv3D(n)                        & bs, w, h, d, n           \\ \bottomrule
\end{tabular}
\label{tab:unet_arch}
\end{table}

\subsection{Baseline Reimplementation Details}

\textbf{LNE.} LNE~\cite{ouyang2021self} is a self-supervised global representation learning method that models longitudinal effects in an autoencoder latent space and is repurposed here for longitudinal segmentation. During pretraining, we require the same base U-Net  architecture for consistency with other baselines. Therefore, to extract a 1024D global representation as in the original paper, we add a global pooling layer on the output of layer $13$ in Table~\ref{tab:unet_arch}, followed by a single 1024D fully-connected layer. With respect to their hyperparameters, due to the larger base network size, we reduce the batch size from 64 to 16 images due to memory considerations. We maintain the other hyperparameters of LNE including downsampling the images to $64^3$, using $N=5$ neighborhoods, and weighing auxiliary losses as $\lambda_{dir}=1$ and $\lambda_{rec}=2$ (see original work for definitions~\cite{ouyang2021self}).  

\textbf{PatchNCE.}
\looseness=-1
PatchNCE~\cite{park2020contrastive} proposes a framework for unsupervised multi-scale patchwise contrastive learning applied to the problem of unpaired image translation and is extended to serve as a baseline in this work as it shares a partially similar motivation. To adapt it to our generic longitudinal representation learning setting, we impose the same modeling assumption as our method (spatial indices in correspondence across an image time-series are positive samples and all other spatial indices are negatives) and modify both its data sampling and loss calculation. First, we perform forward passes on $t$ corresponding intra-subject crops. At each selected layer, $N$ feature vectors are randomly sampled for the longitudinal InfoNCE~\cite{khosla2020supervised} loss:
$\mathcal{L} = \sum_{i}\frac{1}{|P(i)|}\sum_{p\in P(i)} - \log \frac{e^{z_i\cdot z_p/\tau}}{\sum_{s\in S(i)} e^{z_i \cdot z_s/\tau}}$,
where for any query feature index $i$, $P(i)$ is the set of indices of all positives (from the same subject at the same spatial index) and $S(i)$ is the set of all query indices (with size $(t \times N) - 1$, where $t$ is the number of timepoints in the batch and $N$ is the number of patches). We choose $t=3$ timepoints and $N=768$ patches in our experiments due to memory limits.
An ablation study of PatchNCE over the layers used for the loss function and the temperature is given in Table~\ref{tab:patchnce_ablation}. For prototyping efficiency, we perform baseline tuning using a three-layer MLP of size 256 and use validation Dice as the model selection criterion to choose $F$ as our final configuration. For fair comparison with our model, we increase the MLP width in the \texttt{PathNCE} model to 2048 in all other experiments reported, although this does not significantly alter results. 

\textbf{Context Restoration}.
~\cite{chen2019self} proposes context restoration as a pretext task for self-supervised representation learning which consists of randomly and repeatedly swapping image patches and training the network to restore the original image.
We extend Algorithm 1 in ~\cite{chen2019self} to use 3D inputs and targets and tune it with a varying number of swapping iterations (\#swaps) and swapped patch sizes. Downstream validation dice scores are provided in Table~\ref{tab:context_ablation} and configuration $D$ is chosen as the final model.  

\textbf{PCL, GLCL, and 2D U-Net.}
PCL and GLCL~\cite{chaitanya2020contrastive,zeng2021positional} are 2D slice-based contrastive learning methods designed for 3D volume segmentation. We follow the optimal configurations (e.g. temperature, number of partitions, etc.) reported in the original papers. As these methods use a 2D U-Net trained on 2D slices, we additionally benchmark against a randomly initialized 2D U-Net to obtain a baseline for 2D methods. We use a batch size of 128 for all 2D baselines. Within each batch, we compare two different data sampling strategies: (1) randomly sampling 2D slices across all 3D volumes; (2) sampling intra-volume slices within each batch
(i.e. treat one of the dimensions from a $128\times128\times128$ crop as the batch dimension for 2D networks). We train a randomly initialized 2D U-Net for the one-shot segmentation task, and found that (2) significantly increases the validation mean Dice over (1) on OASIS3 from $0.708 (0.193)$ to $0.780 (0.137)$ under the same architecture and optimization set up. We therefore use batch sampling strategy (2) for all 2D model benchmarks (PCL, GLCL, RandInitUnet.2D). 

\begin{table}[h!]
\begin{minipage}[t]{0.45\linewidth}
\centering
\caption{Parameter search (on IBIS-subcort) of patchNCE pretraining over layers included for multiscale patchNCE loss, and temperature. F is used for the final comparison.}
\begin{tabular}{cccc}
\toprule
Exp & layers & temperature & Mean(std) dice      \\ \midrule 
A   & Enc    & 0.07        & 0.834 ($\pm$ 0.060) \\
B   & Enc    & 0.1         & 0.834 ($\pm$ 0.062) \\
C   & Enc    & 0.2         & 0.823 ($\pm$ 0.067) \\
D   & Enc    & 0.5         & 0.830 ($\pm$ 0.064) \\
E   & EncDec & 0.07        & 0.854 ($\pm$ 0.056) \\
F   & EncDec & 0.1         & \textbf{0.856 ($\pm$ 0.054)} \\
G   & EncDec & 0.2         & 0.846 ($\pm$ 0.058) \\
H   & EncDec & 0.5         & 0.850 ($\pm$ 0.054) \\ \bottomrule
\end{tabular}
\label{tab:patchnce_ablation}
\end{minipage}
\hfill
\begin{minipage}[t]{0.45\linewidth}
\centering
\caption{Context restoration parameter search over the size of local patches and the number of patch swap iterations on OASIS3. Validation mean(std) dice is used for selecting the configuration D.}
\begin{tabular}{cccc}
\toprule
Exp & patch size & \# swaps & Mean(std) Dice              \\\midrule
A   & $16^3$         & 10              & 0.778 ($\pm$ 0.141) \\
B   & $24^3$         & 10              & 0.779 ($\pm$ 0.142) \\
C   & $16^3$         & 30              & 0.775 ($\pm$ 0.138) \\
D   & $16^3$         & 50              & \textbf{0.782 ($\pm$ 0.135)}\\ 
E   & $16^3$         & 100             & 0.778 ($\pm$ 0.160) \\\bottomrule
\end{tabular}
\label{tab:context_ablation}
\end{minipage}
\end{table}

\clearpage
\section{Additional data preparation details}\label{app:dataprep}
\subsection{Registration} \label{app:registration}
\subsubsection{Affine registration}
To warp all images to a common space, we warp all images affinely to a constructed template~\cite{avants2010optimal} (with an affine transformation model) using \verb|ANTs|~\footnote{\url{https://github.com/ANTsX/ANTs}} with the following command:
\begin{verbatim}
    antsMultivariateTemplateConstruction2.sh \
    -d 3 \
    -o OUTPUT_FOLDER/T \
    -i 1 -g 0.2 -j 128 -c 2 -r 1 -n 0 -m MI -l 1 \
    -t Affine INPUT_FOLDER/*t1.nii.gz'
\end{verbatim}

Once affinely aligned on a dataset-wide level, we proceed with longitudinal intra-subject deformable alignment for the calculation of $\mathcal{L}_{sim}$ and $\mathcal{L}_{cs}$, as described below.

\subsubsection{Nonlinear Deformable Registration}
Proper spatial correspondence of positive samples for $\mathcal{L}_{sim}$ and label maps for $\mathcal{L}_{cs}$ (the similarity and consistency losses, respectively) requires nonlinear/deformable registration of all intra-subject images to a common reference. To this end, we employ the \verb|ANTs| SyN algorithm from~\cite{avants2008symmetric} to register within-subject images to a single timepoint in a series of acquisitions with the following command:
\begin{verbatim}
    fix = INPUT_FOLDER/{subj}_{trg_tp}_t1.nii.gz
    for src_tp in tps:
        moving = INPUT_FOLDER/{subj}_{src_tp}_t1.nii.gz
        antsRegistration \
        --verbose 1 \
        --dimensionality 3 \
        --float 1 \
        --output [OUTPUT_FOLDER/{subj}_{src_tp}_to_{trg_tp}_t1_, \
                  OUTPUT_FOLDER/{subj}_{src_tp}_to_{trg_tp}_t1_Warped.nii.gz, \
                  OUTPUT_FOLDER/{subj}_{src_tp}_to_{trg_tp}_t1_InvWarped.nii.gz] \
        --transform SyN[0.15, 9, 0.2] \
        --metric CC[{fix}, {moving}, 1, 2, Random, 0.4] \
        --convergence [250x125x50, 1e-5, 10] \
        --shrink-factors 4x2x1 \
        --smoothing-sigmas 2x1x0vox \
        --interpolation Linear
\end{verbatim}
where \texttt{fix} is the intra-subject image from a selected timepoint \texttt{trg\_{tp}} to which all other timepoints \texttt{src\_{tp}} are registered.

\subsection{Additional preprocessing and label generation} \label{app:more_preproc}

\begin{figure}
    \centering
    \includegraphics[width=\textwidth]{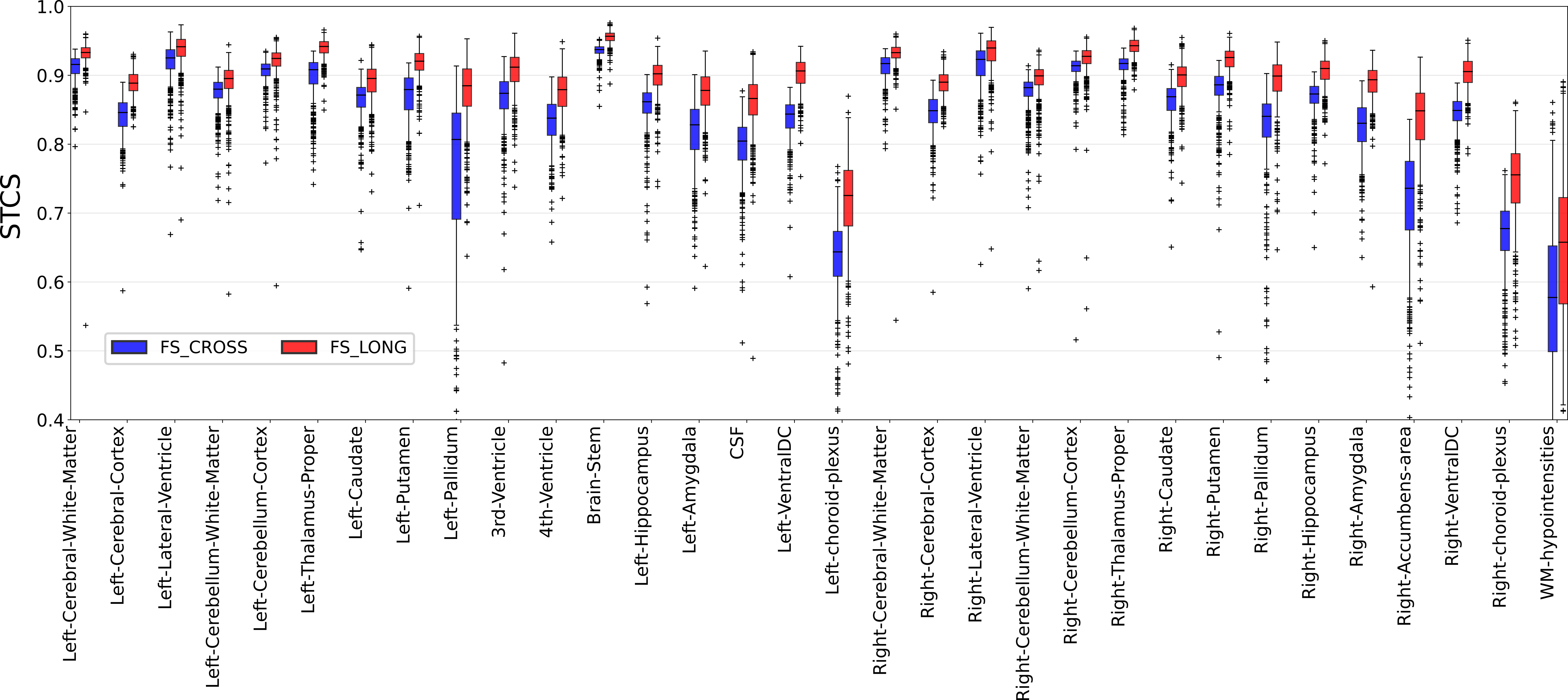}
    \caption{Longitudinal FreeSurfer processing (for anatomical segmentation used as ground-truth targets) leads to substantial improvements over the publicly-released cross-sectional FreeSurfer processing of OASIS3 in longitudinal-consistency metrics such as STCS~\cite{li2021longitudinal} (higher is better). See section \ref{app:more_preproc} for further motivation.}
    \label{fig:fs_cross_vs_long}
\end{figure}

With respect to OASIS3, the publicly-available dataset arrives preprocessed with cross-sectional FreeSurfer processing (intensity and geometric normalization and segmentation). As cross-sectional FreeSurfer does not account for longitudinal effects~\cite{reuter2012within}, FreeSurfer V6.0 is used for additional longitudinal processing on top of the publicly-released cross-sectional FreeSurfer segmentations. A comparison between longitudinal and cross-sectional FreeSurfer analysis based on training set is given in Fig.~\ref{fig:fs_cross_vs_long}, where we observe notable improvements to longitudinal segmentation consistency (as quantified by STCS~\cite{li2021longitudinal}) over all structures. 

For IBIS-\{\textit{subcort, wmgm}\}, all T1w/T2w MR images are preprocessed following standard procedures described in~\cite{shen2022subcortical} of gradient distortion correction, bias-field correction, within-subject and within time-point multi-modality registration, and brain extraction. For IBIS-\textit{subcort}, ground truth label generation follows the segmentation procedures outlined in~\cite{shen2022subcortical}. For IBIS-\textit{wmgm}, we use segmentations generated by a fully-supervised network trained on an external dataset, see Appendix Section \ref{app:wmgm_motivation} for more details.

Once preprocessed and aligned with methods described in Section \ref{app:registration}, the images are cropped to a common field-of-view ($128\times 192\times 160$ for IBIS, $160 \times 192 \times 160$ for OASIS3, corresponding to smaller brain volumes for infants vs. adults) for compatibility with common multi-resolution neural network architectures. 

\subsection{Data splitting with repeat acquisitions}

\setlength{\tabcolsep}{4pt}
\begin{table}[t]
\begin{minipage}{\linewidth}
    \centering
    \includegraphics[width=\textwidth]{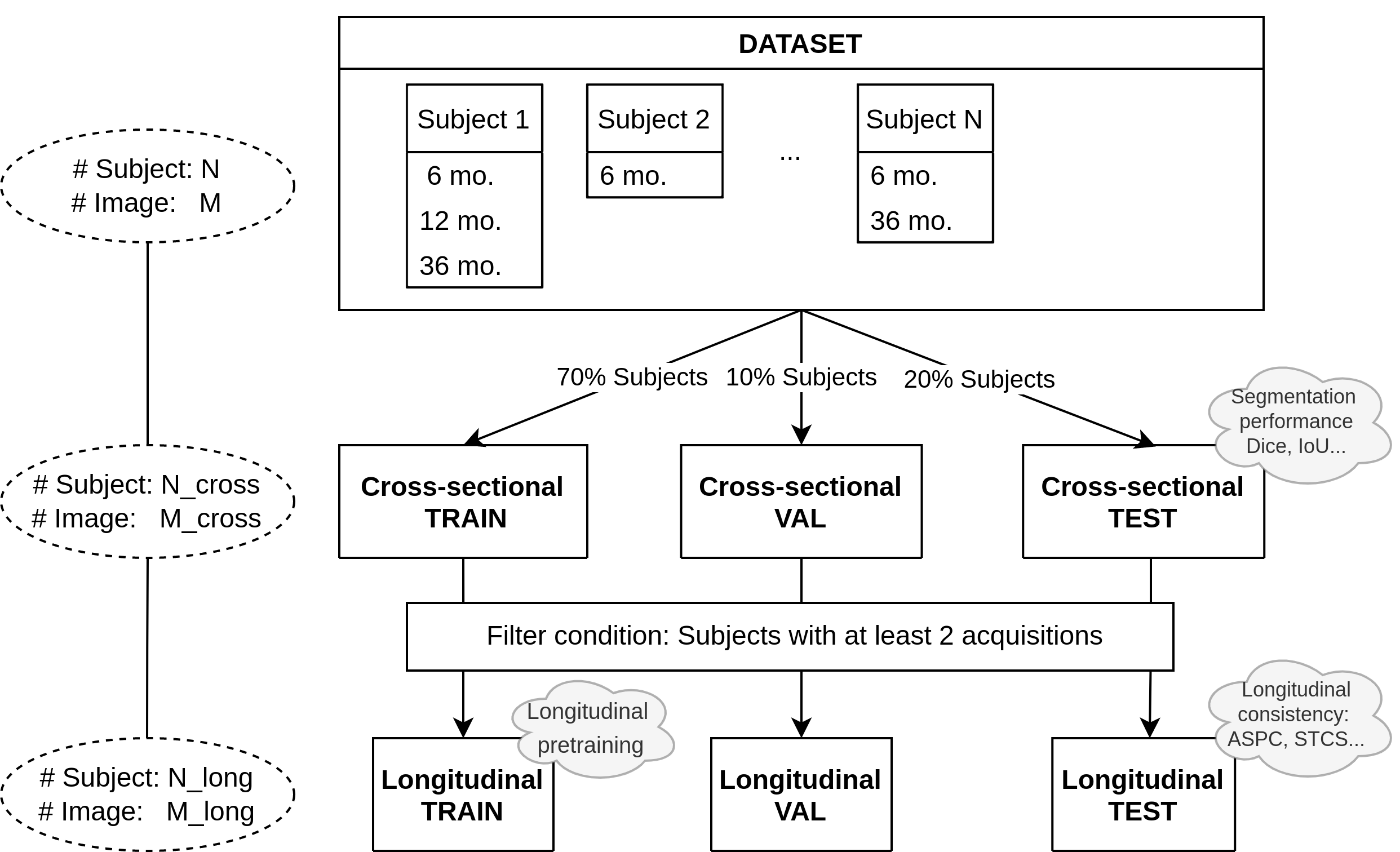}
	\captionof{figure}{\textbf{Longitudinal data selection and splitting criteria.} \textbf{Row 1 $\rightarrow$ 2:} Given $M$ images from $N$ subjects, we apply a 70-10-20 train/val/test \textit{subject-wise} split to avoid data leakage. This level of data split hierarchy is compatible with the calculation of segmentation performance scores which are agnostic to repeat acquisitions such as the Dice coefficient and mean Intersection-over-Union. \textbf{Row 2 $\rightarrow$ 3:} For longitudinal pretraining and calculation of longitudinal consistency scores (such as ASPC and STCS), we further filter the data splits to only contain subjects with two or more acquisitions. Actual dataset-specific numbers reported in Table \ref{tab:N_M_per_dataset}.}
    \label{fig:split}
\end{minipage}
\begin{minipage}{\linewidth}
\centering
\caption{\textbf{Dataset-specific sample sizes of images (N) and subjects (M)}. Table best interpreted in conjunction with Fig. \ref{fig:split}. Abbreviations include \textit{\_long}: longitudinal and \textit{\_cross}: cross-sectional.}
\begin{tabular}{lcccc}
\multicolumn{5}{c}{IBIS}    \\
\toprule
      & \multicolumn{2}{c}{IBIS-subcort}                   & \multicolumn{2}{c}{IBIS-wmgm}                \\ 
 Split      & (N\_cross, M\_cross) & (N\_long, M\_long)    & (N\_cross, M\_cross) & (N\_long, M\_long)  \\ \midrule
Total       & (552, 1272)          & (455, 1175)           &  (552, 1272)  &  (552, 1272)      \\
Train       & (386, 887)           & (313, 814)            &  (386, 887)   &   (313, 814)      \\
Validation   & (55, 133)            & (50, 128)            &  (55, 42)     &    (50, 128)     \\
Test        & (111, 252)           & (92, 233)             &  (111, 80)     &  (92, 233)       \\ \bottomrule \\
\end{tabular}
\begin{tabular}{lcc} 
              \multicolumn{3}{c}{OASIS3}    \\ \toprule
Split       & (N\_cross, M\_cross) & (N\_long, M\_long)   \\ \midrule
Total        & (992, 1639)          & (422, 1069)        \\
Train        & (694, 1147)          & (293, 746)         \\
Validation   & (98, 166)            & (48, 116)          \\
Test         & (200, 326)           & (81, 207)         \\ \bottomrule
\end{tabular}
\label{tab:N_M_per_dataset}
\end{minipage}
\end{table}

We perform a 70-10-20 train, validation, and test split on subject-wise basis from a total of $N$ subjects and $M$ images ($M>N$ due to image time-series acquisitions), with the entire procedure illustrated in Figure \ref{fig:split}. This level of the data split hierarchy  ($N\_cross$ subjects and $M\_cross$ images) is used for segmentation training and evaluation.
For longitudinal pretraining and longitudinal consistency evaluation (ASPC, STCS in Table~\ref{tab:result_moremetrics} and Figure~\ref{fig:result_boxplot} of the main text), we further filter $N\_long$ subjects with at least 2 acquisitions per subject to obtain $M\_long$ images.

Finally, IBIS-\textit{subcort} and IBIS-\textit{wmgm} are segmentation tasks which share the same unlabeled pretraining data. Importantly, we note that as IBIS-\textit{wmgm} is evaluated and tested only on 6-month-old MR images (representing the strongest domain shift, see Appendix Section \ref{app:wmgm_motivation} for its motivation), its segmentation validation and test sets contain fewer images than IBIS-\textit{subcort}. Table \ref{tab:N_M_per_dataset} provides the exact sample sizes for each task.

\clearpage
\section{Miscellaneous Details}\label{app:more_data_details}

\subsection{Data availability and IRB information}\label{app:IRB}
\textbf{IBIS.} The IBIS/Autism MRI data is available through NIH NDA\footnote{\url{https://nda.nih.gov/edit_collection.html?id=19}}. The Infant Brain Imaging Study (IBIS) Network is a National Institutes of Health–funded Autism Center of Excellence project and consists of a consortium of eight universities in the United States and Canada. 
Parents provided informed consent and the institutional review board at each site approved the research protocol.

\textbf{OASIS-3.} OASIS-3~\cite{lamontagne2019oasis} is publicly available through the OASIS webpage\footnote{\url{https://www.oasis-brains.org/}}. Ethical approval was obtained by the relevant ethics committees and informed consent was obtained from all participants following procedures set by the IRB at the Washington University School of Medicine.

\subsection{IBIS-wmgm segmentation task details} \label{app:wmgm_motivation}

\begin{figure}[h]
    \centering
    \includegraphics[width=\textwidth]{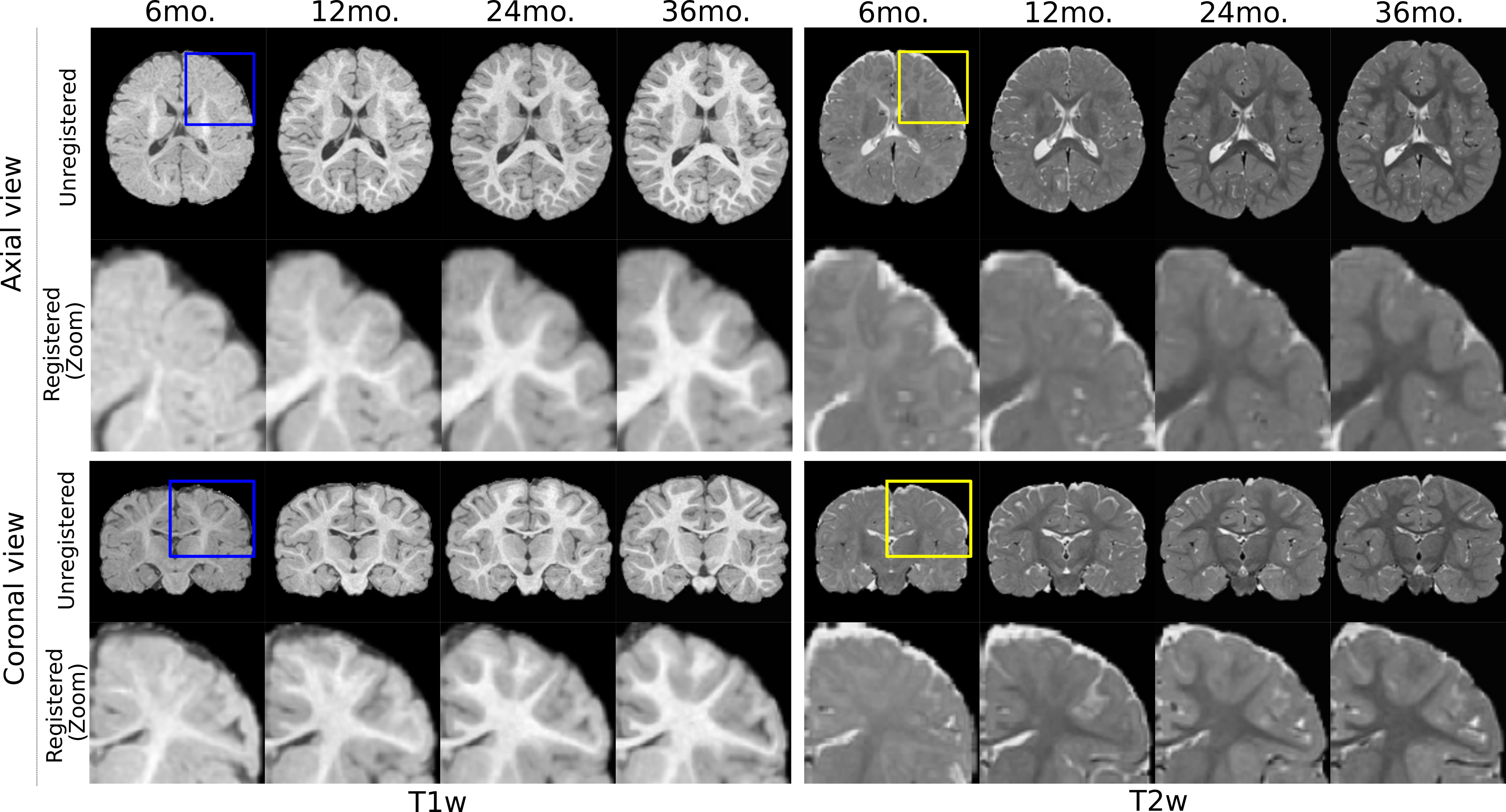}
    \caption{A showcase of \textbf{brain appearance maturation} through the early lifespan. Left: T1w; right: T2w images of developing infant brains from axial/coronal view.}
    \label{fig:domain}
\end{figure}

\textbf{Infant white matter maturation.} Neurodevelopment in infants and toddlers is a highly-complex process of macro and micro-structural changes (for example, growth and myelination, respectively). Relevant to the scope of this paper, real-world tissue segmentation (into grey and white matter) of infant brains at approximately 6 months of age (post-birth) is highly difficult due to ongoing white matter myelination giving the brain an ``isointense" (flat in intensity) appearance which ambiguates anatomical edges. Fig.~\ref{fig:domain} visualizes this phenomenon.

Currently, existing datasets for supervised segmentation of isointense infant brains~\cite{sun2021multi} construct their ground truth labels by first nonlinearly registering algorithmic segmentations of an older timepoint from a given subject image time-series (which is straightforward to segment with existing tools) and then manually correcting the warped labels with the expertise of neuroradiologists.

In our one-shot segmentation setting, we take a partially analogous approach to segmentation of isointense infant brains. We first algorithmically segment~\cite{puonti2016fast} a single arbitrarily-selected 36 month old T1w/T2w MR image and use this reliable segmentation to train all segmentation baselines and the proposed method. These methods which have only been trained on a 36 month old brain are then quantitatively evaluated on isointense 6 month old brain tissue segmentation in this paper. The target labels for the 6 month old evaluation set are generated algorithmically with supervised segmentation networks following~\cite{zeng2018multi} which have been trained on a separate dataset~\cite{sun2021multi}.

\clearpage
\section{Need for regularization} \label{app:needforreg}
\revise{In this section, we qualitatively and quantitatively illustrate that proper regularization of patch-wise \textit{negative-free} multiscale representation learning avoids low-diversity or degenerate decoder representations which hamper downstream segmentation performance. We primarily compare the complete proposed model (\textit{`Ours w/ regularization'}) against an ablation (\textit{`Our ablation w/o regularization'}) that only uses the patch-wise similarity loss and does not use the denoising, variance, covariance, and orthogonality regularization (which corresponds to ablation E from Table~\ref{tab:ablation}).}

\begin{figure}[!hb]
    \centering
    \includegraphics[width=0.90\textwidth]{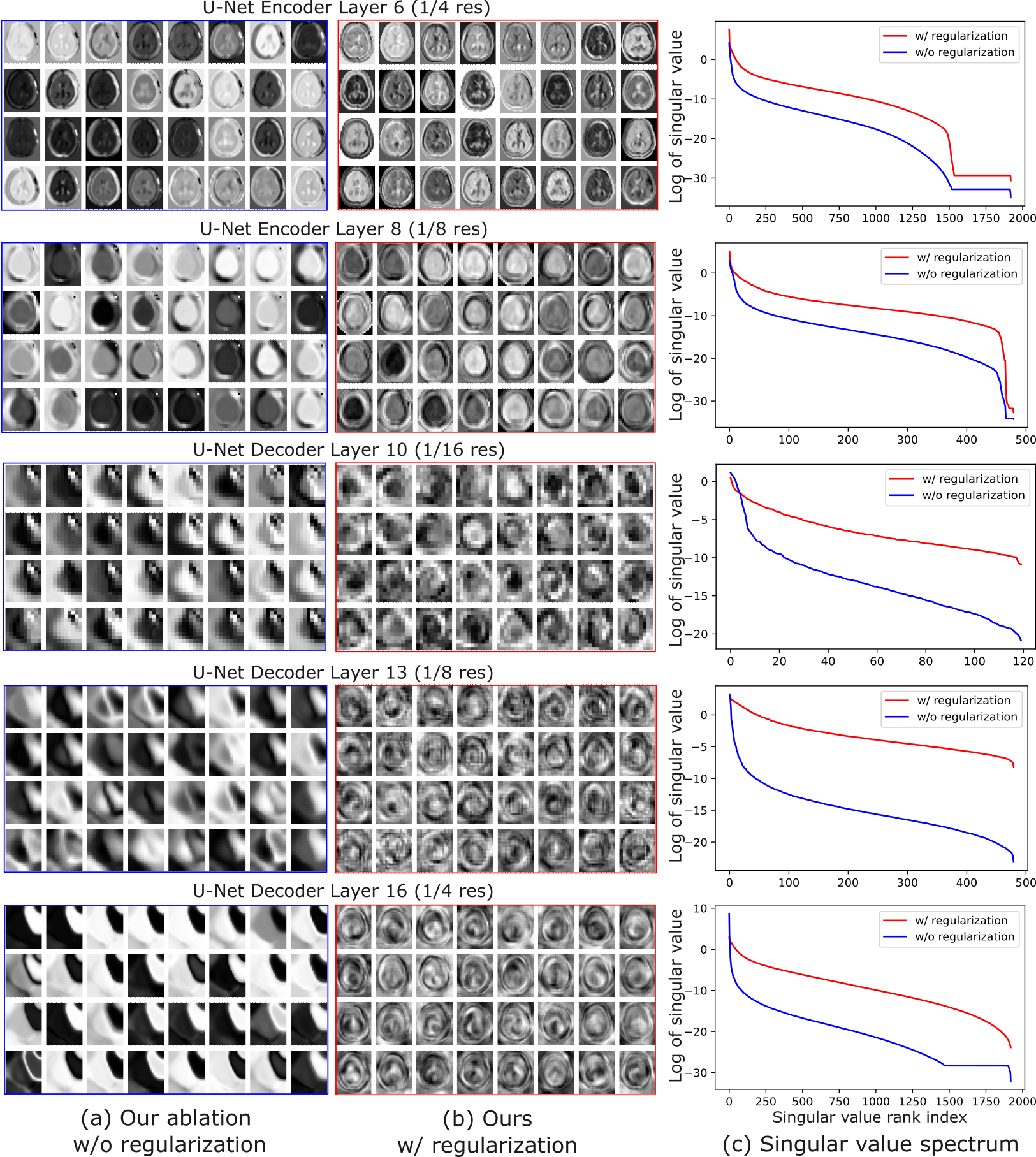}
    \caption{\revise{An illustration of U-Net activations (left) and the log-scale singular values of the spatial covariance matrix $C$ (right), both trained with (red) and without (blue) regularization. While encoder activations are comparable both with and without regularization, the decoder U-Net activations converge to degenerate spatial patterns without regularization (left) and their projections have much lower spatial variability as shown by the singular values of their covariance matrices (right). With regularization, decoder layers avoid degenerate solutions and their spectra indicate higher rank.}}
    \label{fig:collapse_featuremaps_and_svd}
\end{figure}

\looseness=-1
\revise{\textbf{Figure~\ref{fig:collapse_featuremaps_and_svd}} visualizes activations from layers \{6,8,10,13,16\} of Table~\ref{tab:unet_arch} (left) and the sorted singular values of the spatial covariance matrices of multi-layer feature projections from the corresponding layer-wise projector output (right) 
trained with/without regularization. Given a flattened mid-axial feature projection $F \in \mathbb{R}^{wh \times c}$ (where $c = 2048$ and $wh$ is the vectorized spatial dimensionality), we calculate the spatial covariance as $C = \frac{1}{c}\sum_{i=1}^{c} (z_i - \bar{z})(z_i - \bar{z})^{T}$, where $z_i \in \mathbb{R}^{wh \times 1}$ is a spatial feature vector and $\bar{z} = \frac{1}{c} \sum_{i=1}^{c} z_i$. The lower the rank of $C$, the lower the spatial variability of representations.}

\clearpage
\begin{figure}[!ht]
    \centering
    \includegraphics[width=\textwidth]{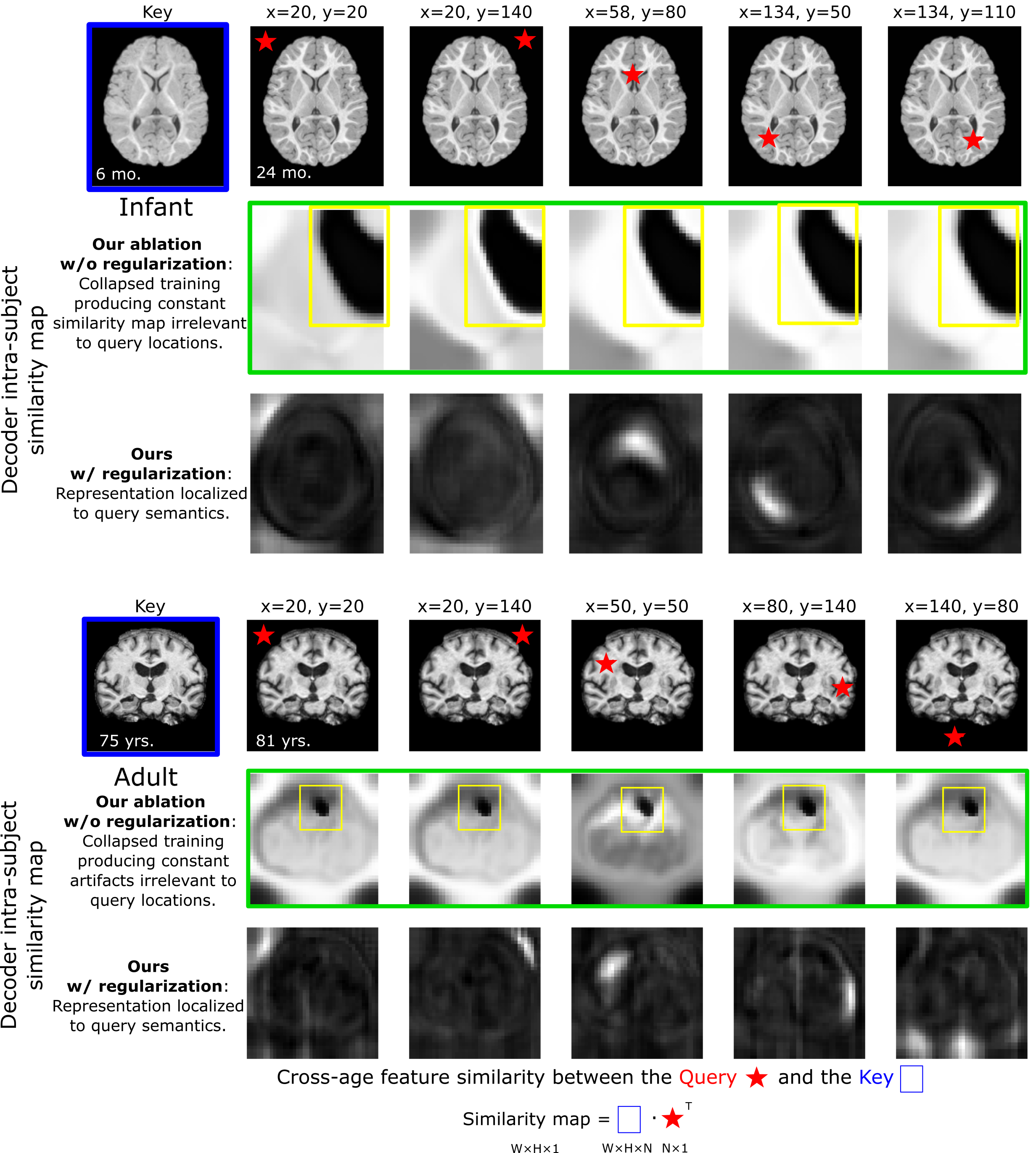}
    \caption{\revise{\textbf{Varying queries against a constant key.} To complement Figures~\ref{fig:simmap} and~\ref{fig:simmap_more}, this figure qualitatively visualizes the intra-subject similarity maps obtained by \textcolor{red}{querying} the projections of decoder layer 16 (from Table~\ref{tab:unet_arch}) from several distinct regions (red stars, columns) and comparing them against the projection of the corresponding decoder activation from a \textcolor{blue}{key} intra-subject temporal image (blue box, left). Without regularization (rows 2 and 5), the ablation yields intra-subject similarity values with very low spatial diversity across all queries, as indicated by similar appearance across all entries in the green boxes and the spatial artifacts in the yellow boxes. However, given proper regularization (rows 3 and 6), these values converge to semantically-similar regions, which indicate better representations for semantically-driven tasks as indicated by improved downstream segmentation performance.}}
    \label{fig:collapse_varysample}
\end{figure}

\clearpage
\begin{figure}
    \centering
    \includegraphics[width=\textwidth]{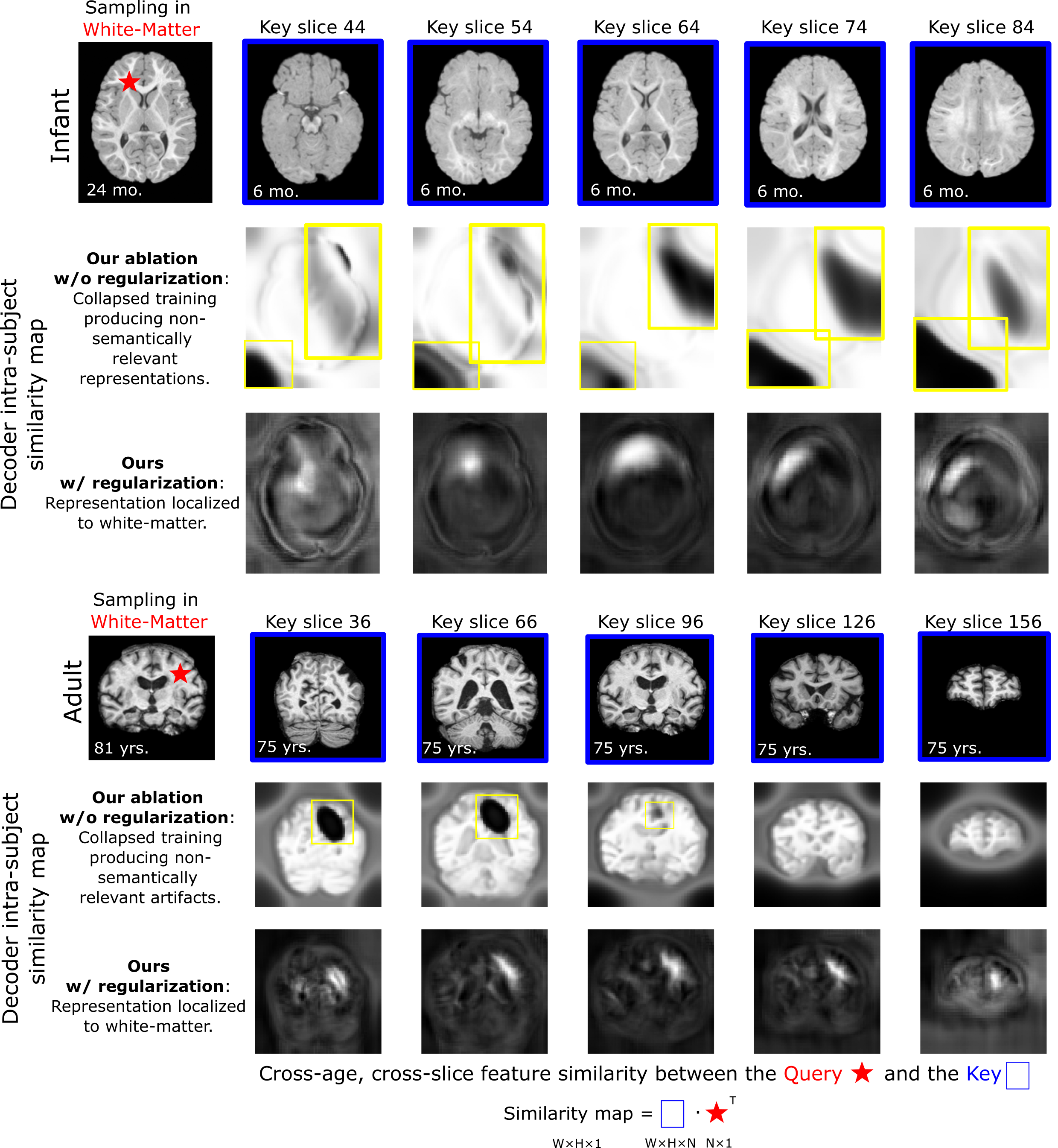} 
    \caption{\revise{\textbf{Varying keys against a constant query.} To complement Figure~\ref{fig:collapse_varysample}, we now hold constant the \textcolor{red}{query} (red star, sampled from white matter) decoder feature projection from layer 19 (of Table~\ref{tab:unet_arch}) and visualize the intra-subject similarity values computed against several distinct \textcolor{blue}{key} slice projections (blue boxes) from the corresponding decoder representations of an intra-subject temporal image. As in Figure~\ref{fig:collapse_varysample}, the ablation without regularization (rows 2 and 5) yields degenerate and semantically-unmeaningful similarity patterns (yellow boxes) with low spatial diversity and artifacts across 3D space. With regularization, self-similarity patterns are semantically-coherent in 3D.}}
    \label{fig:collapse_varyslice}
\end{figure}

\begin{figure}
    \centering
    \includegraphics[width=0.8\textwidth]{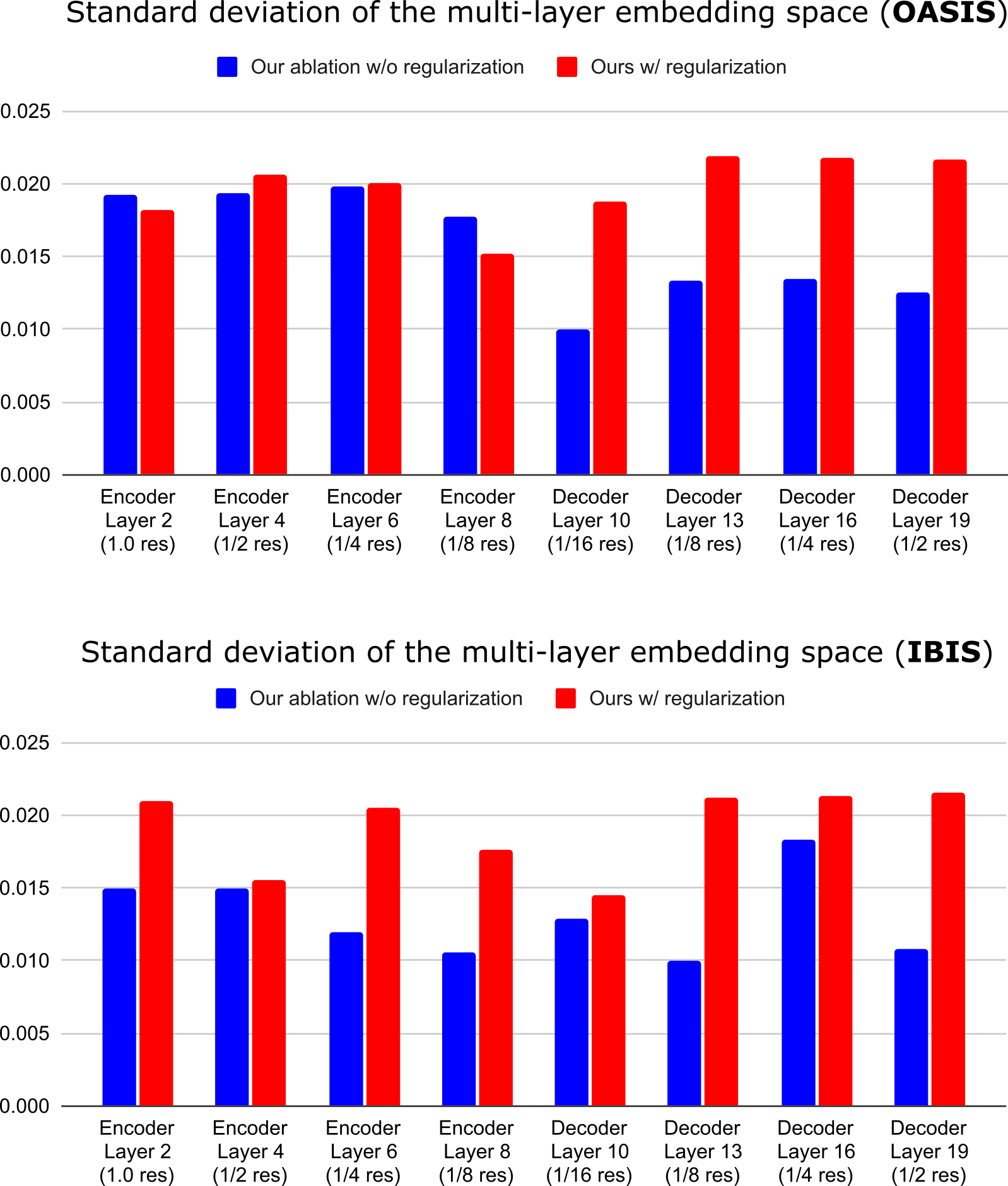}
    \caption{\revise{\textbf{Variability of local U-Net feature embeddings.} This figure visualizes the standard deviations of channel-wise projection vectors from multiple layers of a U-Net pretrained with (red) or without (blue) regularization. As previously suggested by Figures~\ref{fig:collapse_featuremaps_and_svd},~\ref{fig:collapse_varysample}, and~\ref{fig:collapse_varyslice}, decoder projections have significantly lower spatial variability without regularization, indicating low-diversity representations. With regularization, spatial variability is increased which ultimately enables better transfer to downstream tasks such as segmentation.}}
    \label{fig:collapse_std}
\end{figure}

\clearpage
\begin{figure}
    \centering
    \includegraphics[width=\textwidth]{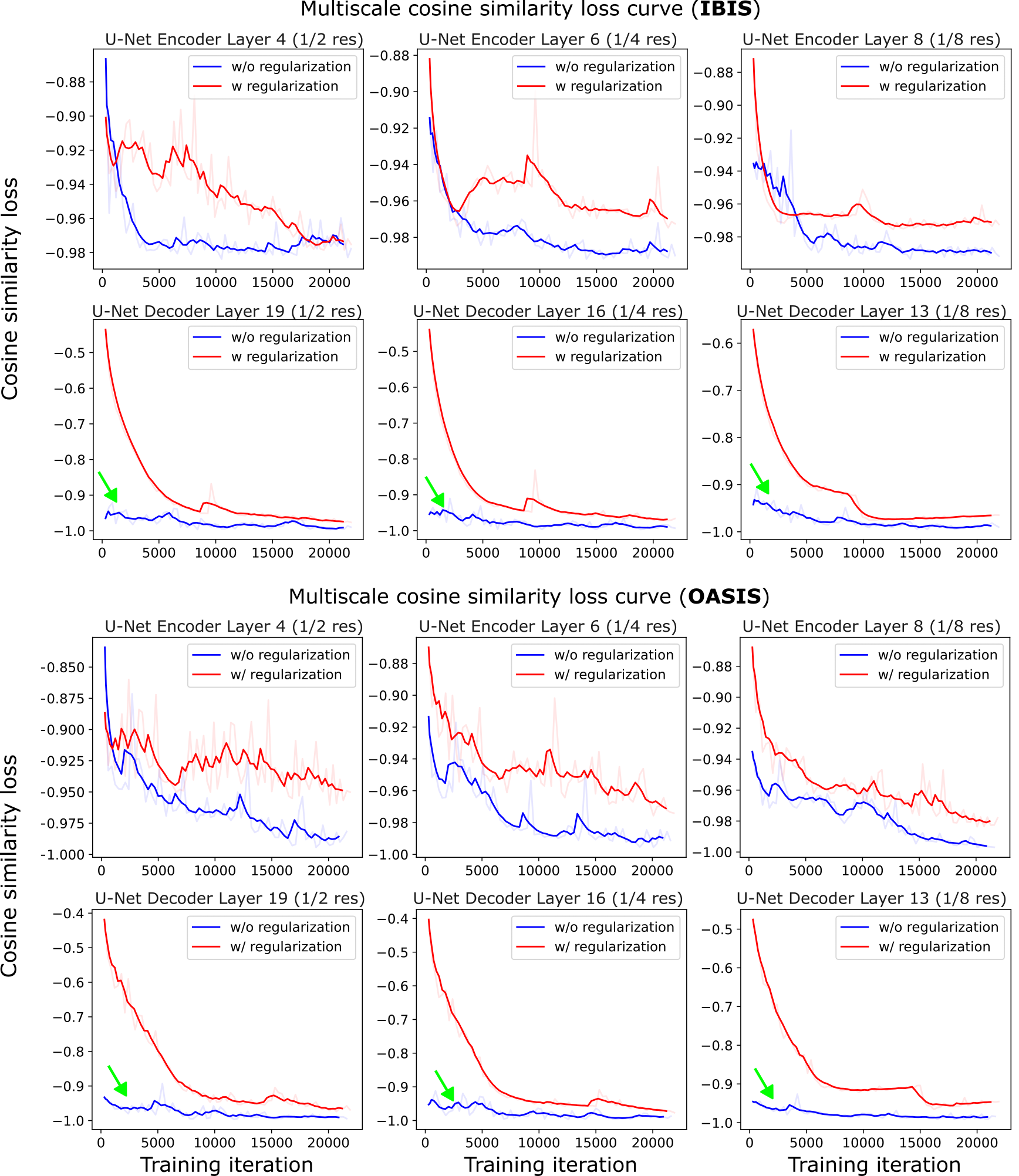}
    \caption{\revise{\textbf{Multi-layer patch-wise cosine similarity loss ($\mathcal{L}_{sim}$).} Here, we visualize the positive-only patch similarity loss against training steps for a selection of encoder and decoder layers (columns) across both datasets with (red) and without (blue) regularization. During initial training stages, we see that encoder layer losses (rows 1 and 3) gradually decrease in both regularized and unregularized settings. However, decoder layer losses (rows 2 and 4) show near-immediate convergence to a degenerate ideal solution (green arrows) without regularization and these unregularized representations do not transfer well to downstream segmentation as shown by the quantitative results in this paper. This phenomenon is consistent with previous literature on negative-free representation learning (see Figure 2a of~\cite{Chen_2021_CVPR}). }}
    \label{fig:collapse_loss}
\end{figure}

\end{document}